\documentclass[runningheads]{llncs}

\usepackage{eccv}

\usepackage{eccvabbrv}

\usepackage{graphicx}
\usepackage{booktabs}
\usepackage[x11names]{xcolor}
\usepackage{colortbl}
\usepackage{multirow}
\usepackage{makecell}
\definecolor{lightblue}{RGB}{218, 232, 252}
\definecolor{lightgray}{RGB}{220,220,220}
\definecolor{lightpurple}{RGB}{230,230,255}
\definecolor{mygray}{gray}{0.92}
\usepackage{wrapfig}
\usepackage[table]{xcolor} 
\usepackage{pifont}        
\newlength\savewidth\newcommand\shline{\noalign{\global\savewidth\arrayrulewidth
  \global\arrayrulewidth 0.8pt}\hline\noalign{\global\arrayrulewidth\savewidth}}

\usepackage[accsupp]{axessibility}  

\newcommand*{\samethanks}[1][\value{footnote}]{\footnotemark[#1]}

\definecolor{eccvblue}{rgb}{0.21,0.49,0.74}
\usepackage[pagebackref,breaklinks,colorlinks,allcolors=eccvblue]{hyperref}

\usepackage{hyperref}

\usepackage{orcidlink}
\usepackage{marvosym}

\begin{document}

\title{Vinci2: Providing Proactive Assistance in Continuous Egocentric Videos} 

\titlerunning{Vinci2}

\author{
Sitong Gong\inst{1,2,3}\orcidlink{0009-0001-8661-1093} \and Tianyu Yan\thanks{Equal contribution.}\inst{1,2,3}\orcidlink{0009-0003-2249-3985} \and
Caixin Kang\samethanks\inst{2,3}\orcidlink{0009-0001-1924-9311} \and 
Bo Zheng\inst{2} \and \\ 
Xiang Ruan\inst{1}\orcidlink{0000-0003-4500-7516} \and
Huchuan Lu\inst{1}\orcidlink{0000-0002-6668-9758} \and
Kaipeng Zhang\inst{2}\orcidlink{0000-0001-6105-6532} \and Yoichi Sato\inst{3}\orcidlink{0000-0003-0097-4537} \and \\
Yifei Huang\textsuperscript{$\dagger$}\inst{2,3}\orcidlink{0000-0001-8067-6227}
}
\authorrunning{Sitong Gong et al.}
\institute{
$^{1}$Dalian University of Technology \quad
$^{2}$Alaya Lab \quad
$^{3}$The University of Tokyo \\
\email{
stgong@mail.dlut.edu.cn, 
\{tianyu.yan, caixin.kang, bo.zheng, kaipeng.zhang\}@shanda.com,
\{ruanxiang, lhchuan\}@dlut.edu.cn,
ysato@iis.u-tokyo.ac.jp, hyf015@gmail.com
}
}

\maketitle

\renewcommand{\thefootnote}{$\dagger$}%
\footnotetext{Corresponding author.}%
\renewcommand{\thefootnote}{\arabic{footnote}}

\begin{abstract}
When should an intelligent assistant speak up without being asked? Continuous egocentric video offers rich, evolving context that enables a new form of assistance: one that is proactive rather than merely reactive. Yet existing approaches either wait passively for user queries or treat every detected event as requiring a response, without considering the user's history, current activity, or whether assistance would actually be welcome. We reframe proactive assistance as a context-dependent decision problem: the agent must not only perceive what is happening, but reason over accumulated temporal context to determine when and whether to intervene. To this end, we present Vinci2, a proactive egocentric assistance system that advances the on-device assistant Vinci from reactive response toward proactivity. On the evaluation side, we present EgoServe, the first large-scale benchmark for proactive assistance in continuous egocentric video. EgoServe comprises over 3,000 service instances organized along 4 temporal memory horizons, ranging from immediate safety alerts to long-term habit coaching, across 10 service categories.
On the modeling side, we propose EgoMemo, a training-free, memory-augmented agent that maintains three complementary memory representations: multi-scale temporal summaries, a semantic knowledge graph, and visual embedding archives. At each timestep, EgoMemo performs retrieval-augmented reasoning to determine whether assistance is warranted and, if so, produces contextually grounded responses. Experiments demonstrate that EgoMemo establishes strong baselines on EgoServe while remaining competitive on existing egocentric benchmarks. Our benchmark and code are publicly available at \href{https://sitonggong.github.io/EgoServe-page/}{Vinci2}. 

\end{abstract}

\begin{figure*}
    \centering
    \includegraphics[width=1\linewidth]{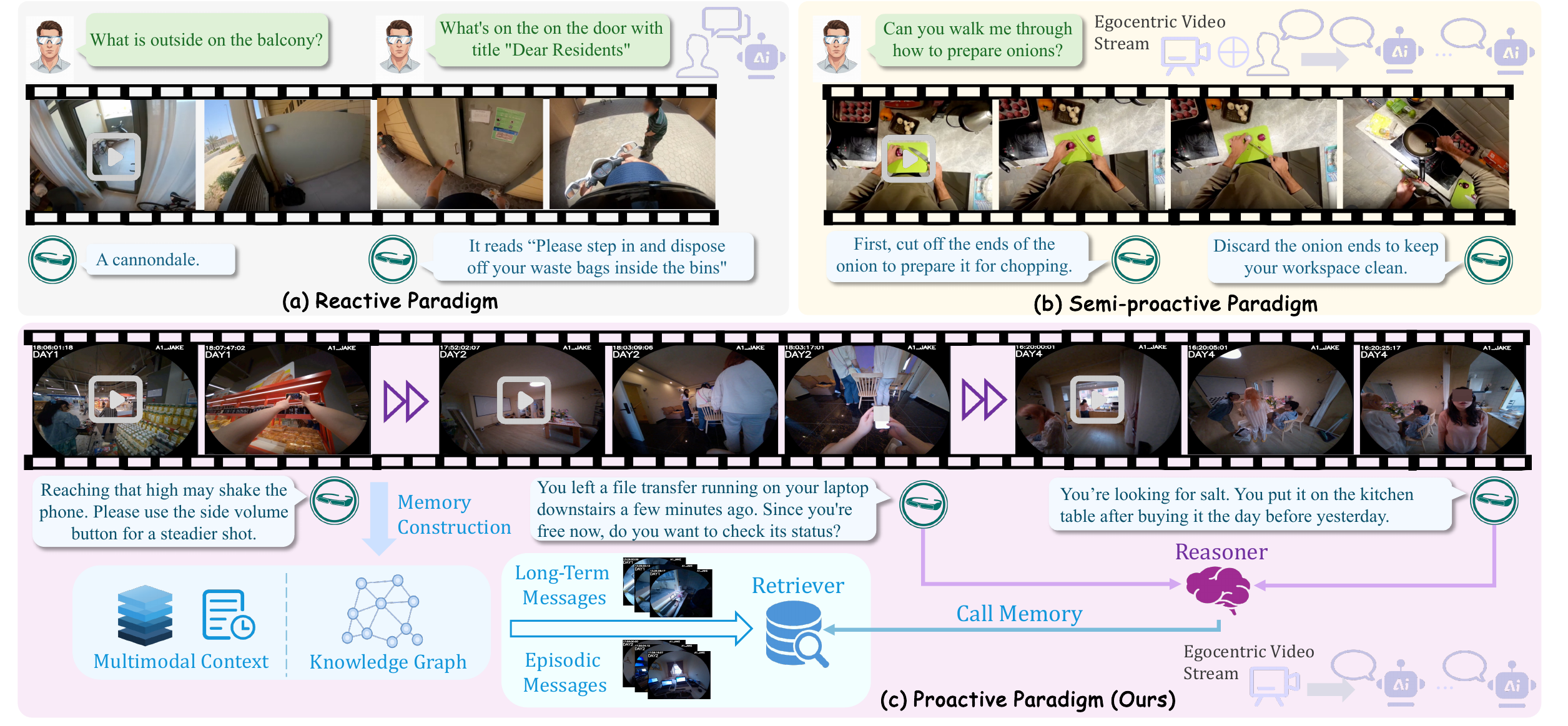}
    \caption{\textbf{Three paradigms of egocentric assistants.} (a) Reactive: responds only to explicit user queries. (b) Semi-proactive: monitors the video stream for predefined task-relevant events. (c) Proactive (ours): autonomously decides when and how to intervene without any user prompt.}
    \label{fig:teaser}
    \vspace{-2em}
\end{figure*}

\section{Introduction}
\label{sec:intro}
The promise of proactive egocentric intelligence is an assistant that sees what you see, understands your context as it evolves, and offers help at the right moment without being asked. Recent progress in Video-LLMs \cite{li2025videochat, lin2024video, shen2024longvu,shu2025video,song2024moviechat}, streaming visual perception \cite{zhang2024flash, qian2025dispider, kang2025can}, and egocentric foundation models \cite{huang2024vinci, yang2025egolife, zhang2024proagent} has brought this vision within reach, enabling continuous comprehension and reasoning over first-person video. Yet while the perception capabilities are maturing rapidly, the question of how and when an assistant should proactively intervene remains largely unaddressed.

Existing egocentric assistants operate under two limiting paradigms, as illustrated in Fig.~\ref{fig:teaser}. Most current Video-LLMs~\cite{li2025videochat, qinghong2022egocentric,zohar2025apollo} follow a \emph{reactive} paradigm, responding only when explicitly prompted. Recent event-triggered systems~\cite{wang2025streambridge, zhang2025eyes, zhang2025proactive} adopt a \emph{semi-proactive} paradigm: given a task instruction provided by the user in advance, they monitor the video stream for predefined events and generate responses upon detection. However, these methods are constrained by the scope of the initial instruction and the immediate visual context, lacking the ability to reason over long-horizon historical observations, and offering no mechanism to assess whether the current situation genuinely warrants interrupting the user. We argue for a third paradigm, \emph{proactive} assistance, that has not yet been explored: the agent reasons over the user's accumulated context, including their history, habits, current activity, and goals, to make a deliberate decision about whether, when, and how to intervene. We instantiate this paradigm in Vinci2, a successor to the egocentric assistant Vinci~\cite{huang2024vinci} that advances from reactive response to genuine proactivity. Vinci2 comprises two complementary components: a benchmark \textbf{EgoServe}, and a training-free agent \textbf{EgoMemo}, which we detail below.

On the evaluation side, no existing benchmark addresses this need. Offline video QA benchmarks~\cite{grauman2022ego4d, mangalam2023egoschema, li2024mvbench, dong2023benchmarking} evaluate comprehension over pre-segmented clips without any notion of proactive intervention. Streaming comprehension benchmarks~\cite{niu2025ovo} assess temporal understanding but do not evaluate proactive behavior. Existing proactive dialogue benchmarks~\cite{wang2025streambridge, zhang2025proactive} operate under a task-completion setting where the user provides an explicit instruction upfront, and do not model the decision of whether to intervene or remain silent. To fill this gap, we present \textbf{EgoServe}, the first large-scale benchmark for proactive assistance in continuous egocentric video. Built upon EgoLife~\cite{yang2025egolife}, HoloAssist~\cite{wang2023holoassist}, and CaptainCook4D~\cite{peddi2024captaincook4d}, EgoServe spans diverse daily activities, multiple users, and extended temporal contexts ranging from minutes to hours, with proactive services organized into 4 temporal memory horizons and 10 service categories.

On the modeling side, current streaming Video-LLMs lack explicit memory mechanisms for long-horizon retrieval, while retrieval-augmented generation methods~\cite{luo2024video} have not been applied to the proactive setting where the system must autonomously decide whether a response is warranted. We take a first step with \textbf{EgoMemo}, a training-free, memory-augmented agent that maintains three complementary memory representations: (1) multi-scale temporal summaries that hierarchically organize observations from fine-grained clip-level descriptions to coarse activity and session summaries; (2) a semantic knowledge graph encoding entity relationships and activity patterns; and (3) visual embedding archives for similarity-based retrieval. At each timestep, the agent determines whether proactive assistance is warranted and, when deeper context is needed, retrieves and synthesizes information across all three memory stores.

Our contributions are as follows:
\begin{itemize}
    \item We formalize proactive assistance in continuous video experiences as a decision-driven reasoning task over streaming egocentric perception, and situate it within a taxonomy of three paradigms: reactive, semi-proactive, and proactive.
    \item We present EgoServe, the first benchmark for evaluating proactive assistance under 4 temporal horizons, comprising over 3,000 service instances across 10 service categories.
    \item We develop EgoMemo, a training-free, memory-augmented agent that demonstrates the feasibility of proactive assistance through retrieval-augmented reasoning, establishing strong baselines on EgoServe and competitive results on existing egocentric benchmarks.
\end{itemize}

\section{Related Works}
\label{sec:related}

\noindent\textbf{Egocentric Video Understanding.}
Egocentric video understanding has been studied primarily in offline settings where complete recordings are available~\cite{girdhar2021anticipative,wang2023ego,plizzari2022e2,radevski2023multimodal,shan2020understanding,zhang2022fine,goyal2022human,huang2018predicting,huang2020mutual,plizzari2023outlook,li2024egoexo,ye2025mmego,huang2020improving,huang2024matching,huang2023weakly}. Benchmarks such as EgoSchema \cite{mangalam2023egoschema}, EgoThink \cite{cheng2024egothink} and EgoExoLearn \cite{huang2024egoexolearn}, EgoExoBench \cite{he2025egoexobench} evaluate episodic memory, cognitive capabilities, and cross-view understanding over pre-recorded footage, while EgoVLP \cite{qinghong2022egocentric} and LaViLa \cite{zhao2023learning} advance video-language pre-training for egocentric retrieval. A parallel line targets the streaming setting: VideoLLM-online \cite{chen2024videollm}, Flash-VStream \cite{zhang2024flash}, and StreamChat \cite{liu2024streamchat} enable real-time video conversation, Dispider \cite{qian2025dispider} disentangles perception and reasoning for low-latency interaction, and OVO-Bench \cite{niu2025ovo} benchmarks online video comprehension. However, both lines focus on answering questions about observed content, without modeling the decision of whether and when to proactively intervene. Our EgoServe fills this gap by evaluating proactive assistance across multiple temporal memory horizons.


\noindent\textbf{Proactive Large Language Models.}
Proactive behavior has been explored in language-only settings, where ProAgent~\cite{zhang2024proagent} enables agents to anticipate teammates' needs in multi-agent cooperation and proactive dialogue systems~\cite{deng2023survey} investigate model-initiated interactions. In the vision-language domain, StreamBridge~\cite{wang2025streambridge}, EWO~\cite{zhang2025eyes}, and ProAssist~\cite{zhang2025proactive} explore event-triggered proactive generation from streaming video, while Vinci~\cite{huang2024vinci} deploys an on-device multimodal proactive assistant. These methods generally conflate event detection with the decision to intervene. In contrast, EgoMemo treats intervention as an explicit reasoning outcome conditioned on retrieved multi-scale historical context.

\noindent\textbf{Proactive Assistive Systems.} The vision of context-aware proactive assistance has deep roots in wearable computing. Early work on context-aware applications \cite{schilit1994context} and contextual awareness in wearable devices \cite{starner1999wearable,dey2001understanding} established the foundational paradigm of systems that adapt behavior based on sensed user context. Recent advances in foundation models have revived this vision: ContextAgent~\cite{yang2025contextagent} builds proactive LLM agents on wearable perceptions from smart glasses and earphones, SensibleAgent \cite{lee2025sensible} introduces unobtrusive proactive interaction for AR glasses, and ProAgentBench \cite{tang2026proagentbench} provides a benchmark for evaluating proactive LLM agents with real-world data. These efforts focus primarily on language-only or short-horizon sensory contexts. In contrast, our work targets proactive assistance over continuous egocentric video streams, requiring long-horizon memory and temporal reasoning across extended activity contexts.

\noindent\textbf{Memory-Based Video Agents.}
Processing long video within limited context windows has motivated memory-augmented architectures such as MovieChat~\cite{song2024moviechat}, MA-LMM~\cite{he2024ma}, and VideoAgent~\cite{fan2024videoagent}. Recent RAG-based approaches~\cite{luo2024video,jeong2025videorag,long2025seeing}, VideoRAG \cite{luo2024video}, Vgent~\cite{shen2025vgent}, and WorldMM~\cite{yeo2025worldmm}, further introduce graph-driven indexing and multi-type memory with adaptive retrieval, but assume offline access to the complete video. EgoMemo departs from these methods in two ways: both memory construction and retrieval are fully streaming, and a VLM-based caption reconstruction step bridges the information gap between structured retrieval results and the contextual descriptions needed for reasoning.

\section{EgoServe Benchmark}

\begin{figure*}
    \centering
    \includegraphics[width=1\linewidth]{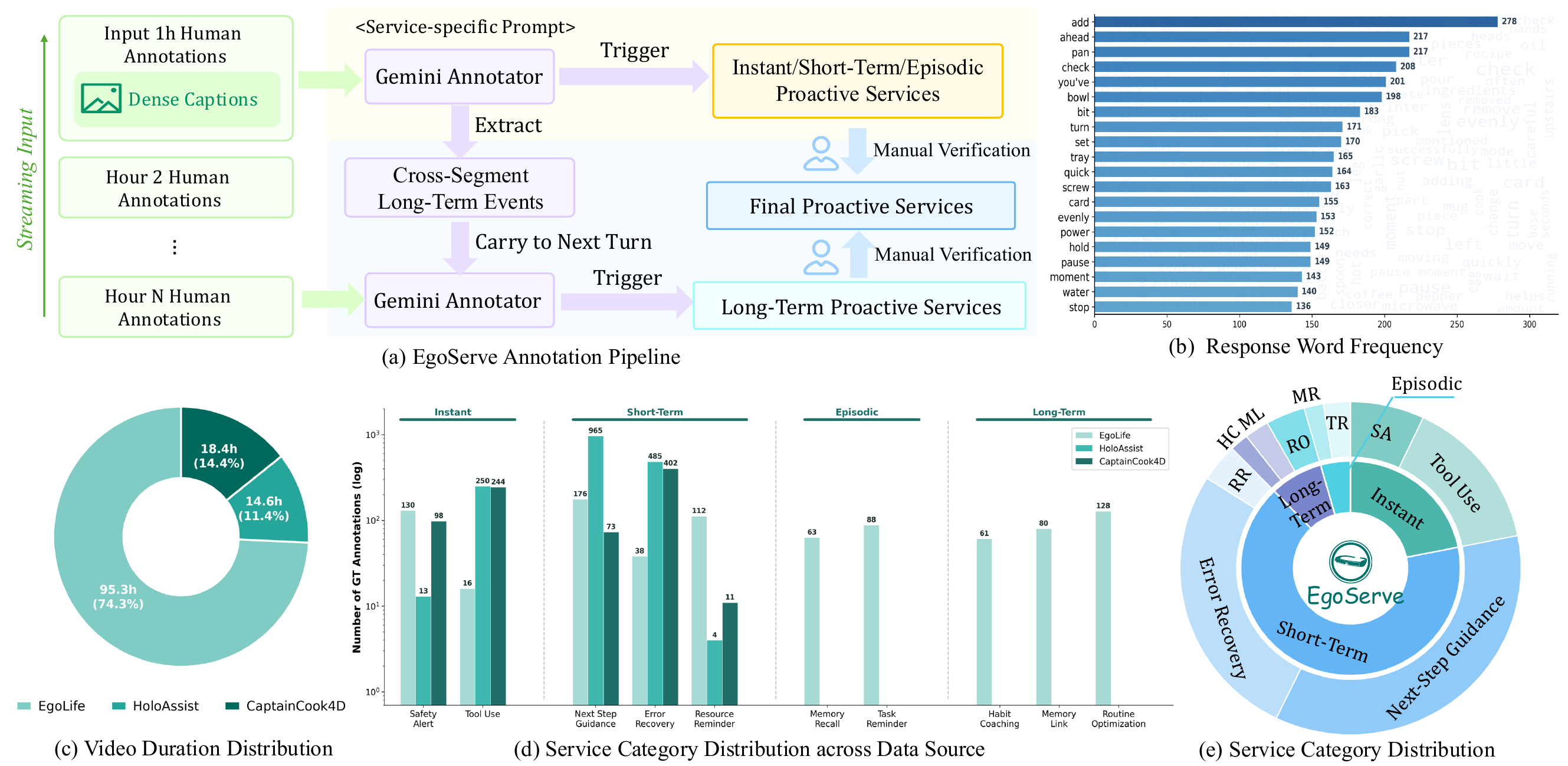}
    \vspace{-2mm}
    \caption{\textbf{Overview of the EgoServe benchmark.} (a) Annotation pipeline: human annotations from each source dataset are processed through category-specific prompts via a foundation model, followed by manual verification. (b) Response word frequency.  (c) Video duration distribution. (d) Per-dataset service counts. (e) Service instance distribution across 10 subcategories and 4 temporal horizons.}
    \vspace{-2em}
    \label{fig:annotation pipeline}
\end{figure*}


\subsection{Task Formulation}
We formulate proactive assistance as a joint decision-and-generation task over continuous video. Let $\mathcal{V} = \{v_1, v_2, \ldots\}$ denote an egocentric video stream segmented into sequential clips. At each timestep 
$t$, the agent observes $v_t$ and must produce a binary intervention decision $d_t \in \{0, 1\}$ along with a service response $r_t$ when $d_t = 1$. A correct proactive response requires three conditions to be met: (1) the intervention occurs within a reasonable temporal window of the ground-truth trigger point; (2) the predicted service type matches the ground-truth category; and (3) the generated response is relevant to the identified service need and grounded in the observed context, as assessed by LLM-based evaluation against reference responses.

\subsection{Data Source}
EgoServe is built upon three egocentric video datasets that together span diverse scenarios and temporal scales. EgoLife~\cite{yang2025egolife} provides multi-day continuous daily life recordings, from which we select three participants (A1, A4, A5) across their first five days, enabling evaluation of long-horizon services that require reasoning across temporally distant events, such as connecting observations from different days.
HoloAssist~\cite{wang2023holoassist} captures task-oriented interactions with procedural annotations (step boundaries, error flags, instructor interventions), and we select its 191 validation videos for instant and short-term service evaluation. CaptainCook4D~\cite{peddi2024captaincook4d} offers structured cooking recordings with both correct and erroneous executions across 24 recipes, from which we select 87 videos with explicit step-error annotations from the validation and test splits, providing a controlled setting for evaluating error detection and corrective guidance.

\subsection{Service Taxonomy}
A key design principle of EgoServe is that proactive services are organized along two orthogonal dimensions: the temporal memory horizon required to provide the service, and the application context of the service itself.
We define 4 temporal memory horizons based on the scope of context the agent must reason over:
\begin{itemize}
    \item \textbf{Instant services} require only the current observation and immediate context, including Safety Alerts (SA: warning about a hazard in the scene) and Tool Use guidance (TU: suggesting a more appropriate tool).
    \item \textbf{Short-Term services} require context spanning the recent minutes of activity, including Error Recovery (ER: detecting and correcting a procedural mistake), Resource Reminder (RR: reminding the user about a recently used resource), and Next-Step Guidance (NSG: suggesting the next action in an ongoing task).
    \item \textbf{Episodic services} require reasoning over the current task, potentially spanning tens of minutes to hours, including Task Reminder (TR: reminding the user of an unfinished task) and Memory Recall (MR: retrieving earlier information that becomes relevant).
    \item \textbf{Long-Term services} require cross-session or multi-day context, including Habit Coaching (HC: suggesting behavioral improvements based on recurring patterns), 
    Routine Optimization (RO: suggesting adjustments to recurring routines), and Memory Link (ML: connecting the current situation to events from previous sessions).
\end{itemize}

This taxonomy yields 4 major categories and 10 subcategories, as summarized in Table~\ref{tab:mainresult}. The design reflects a core insight: the difficulty of proactive assistance scales with the temporal horizon of the required context, and a comprehensive benchmark must evaluate across all horizons.

\subsection{Annotation Pipeline}
\label{annotation pipeline}
Annotating proactive services at scale requires identifying not only what happened in the video, but when assistance would have been appropriate and what the agent should say. We design a semi-automated pipeline that leverages foundation models guided by service category-specific prompts, grounded in existing human annotations from each source dataset.

The annotation pipeline of different datasets is presented in Fig.~\ref{fig:annotation pipeline}. We leverage the existing human annotations and apply different techniques to different data sources. For HoloAssist, we preprocess the full set of human annotations in chronological order and design tailored prompts that map structured annotations into proactive service instances. For example, instructor corrections map to Error Recovery, and step transitions map to Next-Step Guidance. Due to the task-oriented nature of HoloAssist, annotations primarily cover Instant and Short-Term categories.
For EgoLife, we segment annotations into 1-hour intervals and stream them into Gemini together with category-specific prompts. For Instant, Short-Term, and Episodic categories, service dialogues are generated directly from each interval. For Long-Term services, we adopt a streaming cue-capturing strategy: the model incrementally accumulates events across intervals that may trigger specific long-term service types and continuously generates candidate dialogues. These accumulated cues are then combined with future timeline annotations and re-input into the model, simulating the cross-session reasoning that long-term services require.
For CaptainCook4D, we leverage the procedural step annotations and error labels to generate service instances focused on task guidance and error correction.
All generated annotations undergo manual verification to ensure temporal accuracy, category correctness, and response quality.

\subsection{Evaluation Protocol}

EgoServe evaluates proactive assistance along two complementary dimensions:

\noindent\textit{Temporal precision.}
For each service category, we match predicted interventions against ground-truth trigger points using a temporal tolerance window $\delta$ adapted to the characteristic timescale of each source dataset. A prediction is considered a true positive if its trigger timestamp falls within $(\text{start\_time} - \delta,\; \text{end\_time} + \delta)$ of a ground-truth service instance \emph{of the same category}. We compute Precision, Recall, and F1 independently for each of the 10 service subcategories.

\noindent\textit{Response quality.}
For all successfully matched prediction--ground-truth pairs, we evaluate the quality of the generated response using GPT~\cite{openai2025gpt} as an automatic judge, which scores each response on a 1--5 scale across contextual grounding and 
effectiveness. 
The LLM-score reports the average over all matched pairs.

\section{Methodology}
\label{sec:method}

\begin{figure*}[t]
    \centering
    \includegraphics[width=1\linewidth]{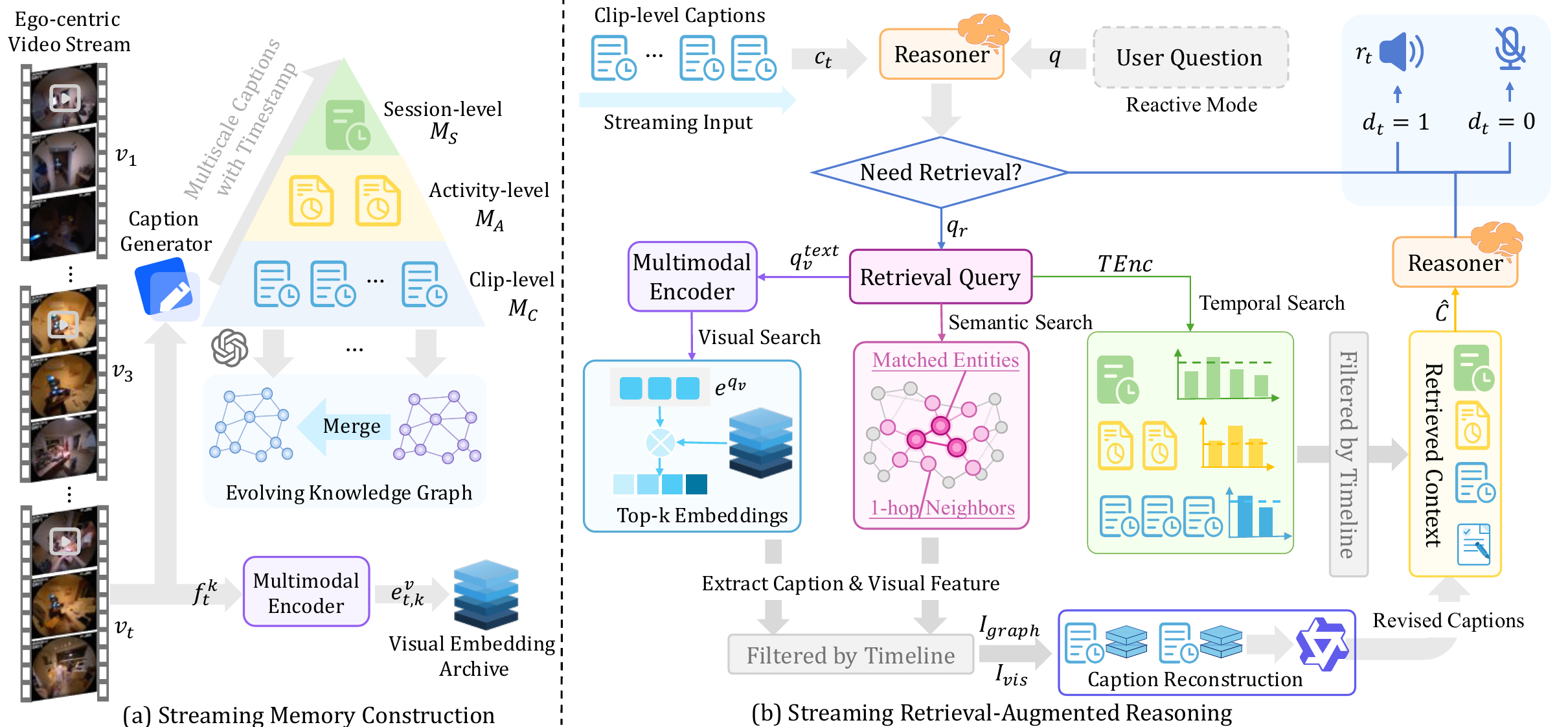}
    \caption{\textbf{Architecture of EgoMemo.} (a) clip-level captions are incrementally organized into three-level temporal summaries ($\mathcal{M}_C$, $\mathcal{M}_A$, $\mathcal{M}_S$), an evolving knowledge graph $\mathcal{G}$, and a visual embedding archive $\mathcal{A}$. (b) Streaming retrieval-augmented reasoning: three parallel retrieval pathways (temporal, semantic, and visual) gather evidence, which is unified via VLM-based caption reconstruction before the reasoner produces an intervention decision $d_t$ and response $r_t$.}
    \vspace{-1em}
    \label{fig:method}
\end{figure*}

We present EgoMemo, a training-free, memory-augmented agent for proactive assistance in continuous egocentric video. As illustrated in Fig.~\ref{fig:method}, EgoMemo continuously processes incoming video, maintains structured long-term memory, and performs context-aware reasoning at each timestep. In \emph{proactive} mode, it monitors the evolving scene and autonomously decides when to provide helpful interventions by retrieving relevant historical context. In \emph{reactive} mode, a user query triggers the same retrieval pipeline. Both modes share a unified architecture consisting of two core stages: (1)~\emph{streaming memory construction} (Sec.~\ref{sec:streamingmemory}), which incrementally builds three complementary memory representations, and (2)~\emph{streaming retrieval-augmented reasoning} (Sec.~\ref{sec:streamingreasoning}), which retrieves and reconstructs relevant context for decision-making. Both construction and reasoning are conducted incrementally, never requiring access to the complete video.

\subsection{Streaming Memory Construction}
\label{sec:streamingmemory}

Given a continuous egocentric video stream $\mathcal{V} = \{v_1, v_2, \ldots\}$, we segment it into non-overlapping short clips. For each clip $v_t$ arriving at timestep $t$, we generate a dense textual caption $c_t$ using a vision-language model, annotated with its corresponding timestamp, and extract sampled keyframes $\{f_t^1, \ldots, f_t^K\}$. These captions and keyframes are the atomic inputs from which three complementary memory representations are incrementally constructed.

\noindent\textbf{Multi-Scale Temporal Memory.}
We organize captions into a three-level hierarchy: \textbf{Clip-level} captions $\mathcal{M}_C = \{(c_t, t) \mid t = 1,\ldots,T\}$ preserve fine-grained perceptual detail; \textbf{Activity-level} summaries $\mathcal{M}_A$ periodically aggregate consecutive clip captions to capture activity context; and \textbf{Session-level} summaries $\mathcal{M}_S$ further aggregate activity entries to encode long-horizon routines. Formally:
\begin{equation}
\label{equa:summarize}
  M_A^{(j)} = \textbf{Summarize}\!\left(\{c_t\}_{t \in \mathcal{W}_j}\right),
  \quad
  M_S^{(k)} = \textbf{Summarize}\!\left(\{M_A^{(j)}\}_{j \in \mathcal{W}_k}\right),
\end{equation}
where $\mathcal{W}_j$ and $\mathcal{W}_k$ denote temporal windows at the Activity and Session levels, respectively. The roll-up is fully incremental: only newly accumulated segments trigger summarization at the next level. To support retrieval, we encode captions at all three levels into dense embeddings $\{e_i^c, e_j^a, e_k^s\}$ using a text encoder \texttt{TEnc} and index them for similarity search.

\noindent\textbf{Evolving Knowledge Graph.}
To capture semantic relationships between entities across different time segments, we maintain a knowledge graph $\mathcal{G} = (\mathcal{N}, \mathcal{E})$ that evolves as new observations arrive. For each caption $c_t$, we prompt an LLM to extract entities and relations $\mathcal{R}_t = \textbf{Extract}(c_t)$, which are merged into the global graph through name-based entity resolution:
$\mathcal{G}_t = \textbf{Merge}(\mathcal{G}_{t-1},\; \mathcal{R}_t).$
Each node in $\mathcal{G}$ maintains links to its source captions, enabling the graph to serve as a structured index over the temporal memory.

\noindent\textbf{Visual Embedding Archive.}
To complement text-based memories with visual details that are difficult to verbalize (object appearances, spatial layouts), we encode sampled keyframes using a multimodal encoder and store the resulting embeddings alongside their source caption index and timestamp:
\begin{equation}
\label{equa:visualembedding}
  \mathbf{e}_{t,k}^v = \textbf{MEnc}(f_t^k) \in \mathbb{R}^{d_v}, \quad
  \mathcal{A} = \{(\mathbf{e}_{t,k}^v,\; t) \mid t = 1,\ldots,T;\; k = 1,\ldots,K\}.
\end{equation}
This enables similarity-based retrieval of visually relevant moments that may lack lexical overlap with the query.


\subsection{Streaming Retrieval-Augmented Reasoning}
\label{sec:streamingreasoning}

At each timestep $t$, the LLM reasoning agent receives the current clip-level caption $c_t$ and the most recent short-term context. For proactive assistance, it assesses whether the current observation warrants an active intervention; for reactive QA, a user query $q$ serves as an external trigger with $d_t = 1$ by default. In both cases, when deeper context is needed, the agent generates a retrieval query $q_r$ and invokes three parallel retrieval pathways, followed by a caption reconstruction step that synthesizes the retrieved evidence into a coherent context for final reasoning.

\noindent\textbf{Multi-Scale Temporal Retrieval.}
We search the multi-scale temporal memory by encoding $q_r$ into a dense embedding and retrieving the top-$k$ similar captions:
\begin{equation}
\label{equa:tempretrieval}
  \mathcal{I}_\text{temp} = \operatorname*{top\text{-}k}_{t'} \operatorname{sim}(\textbf{TEnc}(q_r),\; \mathbf{e}_{t'}^c), \quad t' \le t.
\end{equation}
The constraint $t' \le t$ enforces the streaming setting. Clip-level entries provide fine-grained evidence, while activity- and session-level summaries offer broader context for queries involving extended activities or recurring patterns.

\noindent\textbf{Graph-Based Semantic Retrieval.}
We extract key entities from $q_r$, match them against nodes in $\mathcal{G}$, and expand to their 1-hop neighbors to capture semantically associated events:
\begin{equation}
  \mathcal{N}_\text{match} = \textbf{EntityMatch}(q_r,\; \mathcal{G}), \qquad
  \mathcal{I}_\text{graph} = \bigcup_{n \in \mathcal{N}_\text{match}} \textbf{Neighbor}(n,\; \mathcal{G}),
\end{equation}
where $\mathcal{I}_\text{graph}$ collects caption indices linked to the matched nodes and their neighbors. This captures events described with different wording or occurring at distant timesteps.

\noindent\textbf{Visual Similarity Retrieval.}
Since $q_r$ is in text form, we first rewrite it into a visual-centric description $q_v^\text{text}$ emphasizing visual attributes, encode it via the multimodal encoder to obtain $\mathbf{e}^{q_v} = \textbf{MEnc}(q_v^\text{text})$, and retrieve the top-$k$ matching keyframe embeddings:
\begin{equation}
  \mathcal{I}_\text{vis} = \operatorname*{top\text{-}k}_{t'} \operatorname{sim}(\mathbf{e}^{q_v},\; \mathbf{e}_{t'}^v), \quad t' \le t.
\end{equation}
Each retrieved visual embedding is linked to its source clip-level caption, yielding a set of caption indices $\mathcal{I}_\text{vis}$.

\noindent\textbf{Caption Reconstruction and Reasoning.}
Both graph-based and visual retrieval return sets of caption indices rather than self-contained descriptions. To bridge this gap, we apply a unified \emph{VLM-based caption reconstruction} step: for each set of retrieved indices $\mathcal{I} \in \{\mathcal{I}_\text{graph}, \mathcal{I}_\text{vis}\}$, we gather the corresponding clip-level captions and keyframes, and prompt a VLM to generate a reconstructed caption conditioned on the retrieval query:
\begin{equation}
  \hat{c}_{\mathcal{I}} = \textbf{VLM\_Reconstruct}\!\left(\{(c_{t'}, f_{t'})\}_{t' \in \mathcal{I}},\; q_r\right).
\end{equation}
This reconstruction resolves co-references, recovers visual details absent from the original captions, and produces a query-focused narrative. The outputs from the temporally retrieved captions $\{c_{t'}\}_{t' \in \mathcal{I}_\text{temp}}$, graph-reconstructed context $\hat{c}_\text{graph}$, and visually reconstructed context $\hat{c}_\text{vis}$, are aggregated into a unified retrieved context $\hat{\mathcal{C}}$ and provided to the LLM reasoning agent:
\begin{equation}
\label{equa:reason}
  (d_t, r_t) = \textbf{Reason}(c_t,\; \hat{\mathcal{C}},\; q_r),
\end{equation}
where $d_t \in \{0,1\}$ is the intervention decision and $r_t$ is the generated response. For reactive QA, the same pipeline is invoked with the user query as $q_r$ and $d_t = 1$. This unified formulation requires no architectural modification between the two modes.

\subsection{Adaptation to Other Benchmarks}
\label{sec:offline}

While EgoMemo is designed for streaming scenarios, its architecture naturally accommodates offline video understanding tasks where the full video is available at inference time. In this setting, we first process the entire video sequentially to construct the complete memory representations ($\mathcal{M}_C$, $\mathcal{M}_A$, $\mathcal{M}_S$, $\mathcal{G}$, and $\mathcal{A}$) in a single forward pass. Given a query $q$, the agent adopts a coarse-to-fine perception strategy: it first receives all activity-level and session-level summaries ($\mathcal{M}_A$ and $\mathcal{M}_S$) to form a global understanding of the video, and attempts to answer the question based on this high-level context alone. If the agent determines that the available information is insufficient, it invokes the same retrieval pipeline described in Sec.~\ref{sec:streamingreasoning} to retrieve fine-grained clip-level evidence from $\mathcal{M}_C$, $\mathcal{G}$, and $\mathcal{A}$. This coarse-to-fine strategy avoids unnecessary retrieval for questions answerable from high-level summaries, while preserving access to detailed evidence when needed. The adaptation requires no modification to the model architecture, demonstrating that the streaming-first design of EgoMemo generalizes gracefully to offline settings. We evaluate both streaming and offline performance in Sec.~\ref{sec:experiments}.

\section{Experiments}
\label{sec:experiments}
\subsection{Experimental Setup}

\subsubsection{Evaluation Datasets}
We evaluate EgoMemo across three benchmark categories. (1)~\textbf{Proactive assistance}: our EgoServe benchmark, which evaluates proactive service triggering over streaming egocentric video. (2)~\textbf{Online video understanding}: ESTP-Bench~\cite{zhang2025eyes}, which assesses ego-proactive video understanding through temporally grounded questions at opportune moments; and OVO-Bench~\cite{niu2025ovo}, which evaluates online video comprehension across backward tracing, real-time perception over 858 videos. (3)~\textbf{Offline egocentric QA}: EgoSchema~\cite{mangalam2023egoschema}, a long-form temporal reasoning benchmark with over 5,000 multiple-choice questions spanning 250+ hours of Ego4D video; EgoTaskQA~\cite{jia2022egotaskqa}, which tests understanding of task-oriented egocentric activities, converted to multiple-choice format following~\cite{fan2024videoagent}; and QAEgo4D~\cite{barmann2022did}, which evaluates episodic memory querying over Ego4D recordings.

\subsubsection{Implementation Details} For streaming memory construction, we use Qwen3-VL-8B-Instruct as the VLM to generate clip-level captions $\mathcal{M}_L$ from each video segment. Activity-level summaries $\mathcal{M}_A$ and Session-level summaries $\mathcal{M}_S$ are produced via LLM-based summarization. The temporal windows $\mathcal{W}_j$ and $\mathcal{W}_k$ are adapted to the video duration: for ultra-long recordings such as EgoLife, we use 30s/5min/1h for clip/activity/session levels, while for shorter online benchmarks we use 10s/1min/5min. Entity and relation extraction for the evolving knowledge graph is performed using GPT-4o-mini. For retrieval, we encode textual captions with OpenAI's text-embedding-3-small as the text encoder \texttt{TEnc}, and encode keyframes with ImageBind as the multimodal encoder \texttt{MEnc}. The reasoning agent uses GPT-5-mini for online streaming benchmarks and GPT-5.2 for offline benchmarks. For EgoServe evaluation, the temporal tolerance window $\delta$ is set per dataset according to its characteristic timescale: $\delta=60$s for EgoLife, $\delta=25$s for CaptainCook4D, and $\delta=10$s for HoloAssist. Further details on hyperparameters and prompt designs are provided in the supplementary material.

\renewcommand{\arraystretch}{1.2}
\begin{table*}[t]
\centering
\caption{Evaluation results on EgoServe benchmark.}
\label{tab:mainresult}
\small
\setlength{\tabcolsep}{3pt}
\resizebox{1.0\textwidth}{!}{
\begin{tabular}{l | cc | ccc | cc | ccc | c | c}
\Xhline{1.0pt}
\rowcolor{mygray}
  & \multicolumn{2}{c|}{\textbf{Instant}}
  & \multicolumn{3}{c|}{\textbf{Short-term}}
  & \multicolumn{2}{c|}{\textbf{Episodic}}
  & \multicolumn{3}{c|}{\textbf{Long-term}}
  & & \\
\rowcolor{mygray}
\multirow{-2}{*}{\textbf{Model}}
  & \textit{SA} & \textit{TU}
  & \textit{NSG} & \textit{ER} & \textit{RR}
  & \textit{MR} & \textit{TR}
  & \textit{HC} & \textit{ML} & \textit{RO}
  & \multirow{-2}{*}{\textbf{Overall}}
  & \multirow{-2}{*}{\textbf{LLM-score}} \\
\hline
Qwen3-VL-Plus~\cite{bai2025qwen3}
  & 5.2 & 1.5 & 8.6 & \textbf{10.4} & 3.4 & 0.0 & 1.8 & 4.4 & 0.0 & 0.0 & 3.5 & 2.8 \\
GPT-5-mini~\cite{openai2025gpt} & 
\textbf{12.5} & 3.6 & 9.5 & 1.0 & 5.2 & 0.0 & \textbf{9.4} & \textbf{5.7} & 0.0 & 0.0 & 4.7 & \textbf{3.0} \\
\hline
w/o MS
& 11.2 & 8.0 & 24.2 & 1.9 & 4.9 & \textbf{4.8} & 2.9 & 4.6 & 1.9 & 5.7 & 7.0 & 2.8 \\
w/o Recons.
& 10.5 & 7.6 & \textbf{26.1} & 0.4 & \textbf{5.8} & 0.0 & 4.8 & 5.4 & 0.0 & 5.7 & 6.6 & 2.8 \\
w/o VSR
& 11.7 & 6.0 & 24.8 & 1.2 & 3.9 & 1.2 & 6.6 & 4.2 & 3.2 & 6.4 & 6.9 & 2.8 \\
w/o GSR & 
\textbf{12.5} & 5.6 & 24.4 & 1.9 & 4.9 & 2.6 & 4.8 & 2.4 & 0.0 & 5.4 & 6.5 & 2.8  \\
w/o MTR & 
9.7 & \textbf{8.8} & 24.6 & 1.2 & 4.0 & 4.0 & 4.5 & 1.9 & 2.1 & 7.5 & 6.8 & 2.7\\
\Xhline{0.5pt}
\rowcolor{lightpurple}
\textbf{EgoMemo (Ours)} & 
11.4 & 7.5 & 24.7 & 1.7 & 4.7 & 3.8 & 5.7 & 3.7 & \textbf{4.9} & \textbf{11.8} & \textbf{8.0} & 2.8 \\
\Xhline{1.0pt}
\end{tabular}
\vspace{-5mm}
}
\end{table*}
\vspace{-5mm}
\renewcommand{\arraystretch}{1.0}

\subsection{Results on EgoServe}
\subsubsection{Main Comparison}
Table~\ref{tab:mainresult} reports per-category F1 scores on the EgoServe benchmark, where a prediction counts as a true positive only if it falls within the dataset-specific temporal window of a ground-truth instance of the \emph{same} service category. 
We compare EgoMemo against two strong proprietary models: GPT-5-mini and Qwen3-VL-Plus. 
Both models struggle substantially: 
Qwen fails almost entirely on episodic and long-term services where historical context is essential. GPT-5-mini performs better (4.7 overall), particularly on instant Safety alerts (12.5) and short-term Next-Step Guidance (9.5), but still scores zero on all long-term categories except Habit Coaching (5.7), confirming that even powerful VLMs cannot provide meaningful proactive assistance without access to accumulated context.

EgoMemo achieves 8.0 overall F1, nearly doubling GPT-5-mini's performance. The gains are most pronounced on long-term services: Memory Link reaches 4.9 (vs.\ 0.0 for both baselines) and Routine Optimization reaches 11.8 (vs.\ 0.0), demonstrating that our structured memory and retrieval mechanisms successfully enable cross-session reasoning. On episodic services, EgoMemo attains non-zero scores on both Memory Recall (3.8) and Task Reminder (5.7), whereas the baselines fail on at least one. Short-term services show strong performance across Next-Step Guidance (24.7) and Resource Reminder (4.7). For matched predictions, the LLM-score reaches 2.8. We note that GPT-5-mini matches far fewer service instances overall (4.7 vs.\ 8.0), indicating that the primary bottleneck for baselines lies in temporal precision rather than response quality.

The absolute scores remain moderate across all methods, reflecting the genuine difficulty of proactive assistance: the agent must simultaneously decide \emph{when} to intervene within the correct temporal window, identify the appropriate service type, and produce a helpful response. EgoServe is designed to expose these challenges and serve as a diagnostic tool for future progress.

\renewcommand{\arraystretch}{1.2}
\begin{table*}[t]
\centering
\caption{Experimental results of various models evaluated on the ESTP-Bench.}
\label{tab:ESTP_result}
\vspace{-2mm}
\small
\setlength{\tabcolsep}{3pt}
\resizebox{1.0\linewidth}{!}{
\begin{tabular}{l | ccccccccc | ccccc}
\Xhline{1.0pt}
\rowcolor{mygray}
  & \multicolumn{9}{c|}{\textbf{Explicit Proactive Task}}
  & \multicolumn{5}{c}{\textbf{Implicit Proactive Task}} \\
\rowcolor{mygray}
\multirow{-2}{*}{\textbf{Model}} & \textit{OR} & \textit{AP} & \textit{TRU} & \textit{OL} & \textit{OSC}
 & \textit{EOL} & \textit{EOSC} & \textit{AR} & \textit{All}
 & \textit{OFR} & \textit{IFR} & \textit{NAR} & \textit{TU} & \textit{All} \\
\hline
\textcolor{gray}{EyeWO*}
  & \textcolor{gray}{26.6} & \textcolor{gray}{26.6} & \textcolor{gray}{25.1}
  & \textcolor{gray}{26.8} & \textcolor{gray}{19.8} & \textcolor{gray}{22.3}
  & \textcolor{gray}{20.8} & \textcolor{gray}{20.7} & \textcolor{gray}{23.6}
  & \textcolor{gray}{24.8} & \textcolor{gray}{31.0} & \textcolor{gray}{75.3}
  & \textcolor{gray}{78.7} & \textcolor{gray}{52.5} \\
\hline
LLaVA-OneVision~\cite{li2024llava}
  & 8.3  & 8.8  & 22.8 & 25.4 & 13.5
  & 9.8  & 9.6  & 10.3 & 13.6
  & 20.3 & 20.9 & 35.9 & 49.9 & 31.8 \\
Qwen2-VL~\cite{bai2023qwen}
  & 13.7 & 13.5 & 15.4 & \textbf{29.5} & 8.0
  & 15.4 & 16.6 & 10.9 & 15.4
  & 17.8 & 19.8 & \textbf{56.4} & 63.1 & \textbf{39.3} \\
MiniCPM-V~\cite{yao2024minicpm}
  & 14.9 & 16.8 & 17.1 & 26.8 & 7.7
  & 12.9 & 12.5 & 13.1 & 15.2
  & 15.9 & 21.0 & 46.8 & 62.2 & 36.5 \\
LLaVA-NeXT-Video~\cite{li2024llava}
  & 15.6 & 14.6 & 21.9 & 26.8 & 12.8
  & 14.2 & 13.5 & 12.3 & 16.5
  & 18.6 & 23.2 & 44.9 & 51.6 & 34.6 \\
InternVL-V2~\cite{chen2024expanding}
  & 11.3 & 5.9  & 7.0  & 10.1 & 0.7
  & 2.7  & 5.2  & 2.2  & 5.6
  & 8.3  & 2.9  & 4.3  & 11.2 & 6.7  \\
LIVE~\cite{chen2024videollm}
  & 11.2 & 13.9 & 7.9  & 13.2 & 5.6
  & 9.4  & 6.0  & 8.9  & 9.5
  & 5.8  & 8.9  & 41.0 & 46.7 & 25.6 \\
MMDuet~\cite{wang2024videollm}
  & 7.2  & 10.3 & 17.6 & 10.2 & 4.2
  & 6.1  & 8.8  & 8.5  & 9.1
  & 10.0 & 7.7  & 50.1 & \textbf{69.1} & 34.2 \\
\Xhline{0.5pt}
\rowcolor{lightpurple}
\textbf{EgoMemo (Ours)}
  & \textbf{25.4} & \textbf{30.5} & \textbf{32.4} & 22.9 & \textbf{26.6}
  & \textbf{26.1} & \textbf{35.8} & \textbf{21.1} & \textbf{27.6}
  & \textbf{23.8} & \textbf{29.9} & 36.7 & 48.4 & 34.7 \\
\Xhline{1.0pt}
\end{tabular}
}
\vspace{-0.2cm}
\end{table*}
\renewcommand{\arraystretch}{1}

\renewcommand{\arraystretch}{1.2}
\begin{table*}[h]
\centering
\caption{Evaluation results on OVO-Bench.}
\vspace{-2mm}
\label{tab:ovobench}
\small
\setlength{\tabcolsep}{2pt}
\resizebox{1.0\textwidth}{!}{
\begin{tabular}{l | ccccc c | c | cc c | c}
\Xhline{1.0pt}
\rowcolor{mygray}
  & \multicolumn{7}{c|}{\textbf{Real-Time Visual Perception}}
  & \multicolumn{4}{c}{\textbf{Backward Tracing}} \\
\rowcolor{mygray}
\multirow{-2}{*}{\textbf{Model}}
  & \textit{OCR} & \textit{ACR} & \textit{ATR} & \textit{STU} & \textit{FPD} & \textit{OJR} & \textit{Avg.}
  & \textit{EPM} & \textit{ASI} & \textit{HLD} & \textit{Avg.} \\
\hline
GPT-4o~\cite{gpt4o}
  & 69.8  & 64.22 & 71.55 & 51.12 & 70.3  & 59.78 & 64.46
  & \textbf{57.91} & \textbf{75.68} & 48.66 & \textbf{60.75} \\
Qwen2-VL-72B~\cite{bai2023qwen}
  & 65.77 & 60.55 & 69.83 & 51.69 & 69.31 & 54.35 & 61.92
  & 52.53 & 60.81 & \textbf{57.53} & 56.95 \\
LLaVA-Video-7B~\cite{li2024llava}
  & 69.13 & 58.72 & 68.83 & 49.44 & \textbf{74.26} & 59.78 & 63.52
  & 56.23 & 57.43 & 7.53  & 40.4  \\
LLaVA-OneVision-7B~\cite{li2024llava}
  & 66.44 & 57.80 & 73.28 & 53.37 & 71.29 & 61.96 & 64.02
  & 54.21 & 55.41 & 21.51 & 43.71 \\
InternVL-V2-8B~\cite{chen2024expanding}
  & 67.11 & 60.55 & 63.79 & 46.07 & 68.32 & 56.52 & 60.39
  & 48.15 & 57.43 & 24.73 & 43.44 \\
LongVU-7B~\cite{shen2024longvu}
  & 53.69 & 53.21 & 62.93 & 47.75 & 68.32 & 59.78 & 57.61
  & 40.74 & 59.46 & 4.84  & 35.01 \\
VideoLLM-online-8B~\cite{chen2024videollm}
  & 8.05  & 23.85 & 12.07 & 14.04 & 45.54 & 21.20 & 20.79
  & 22.22 & 18.80 & 12.18 & 17.73 \\
\textcolor{black}{EyeWO~\cite{zhang2025eyes}}
  & \textcolor{black}{24.16} & \textcolor{black}{27.52} & \textcolor{black}{31.89} & \textcolor{black}{32.58} & \textcolor{black}{44.55} & \textcolor{black}{35.87} & \textcolor{black}{32.76}
  & \textcolor{black}{39.06} & \textcolor{black}{38.51} & \textcolor{black}{6.45}  & \textcolor{black}{28.00} \\
Dispider~\cite{qian2025dispider}
  & 57.72 & 49.54 & 62.07 & 44.94 & 61.39 & 51.63 & 54.55
  & 48.48 & 55.41 & 4.3   & 36.06 \\
\Xhline{0.5pt}
\rowcolor{lightpurple}
\textbf{EgoMemo (Ours)}
  & \textbf{91.28} & \textbf{75.23} & \textbf{77.59} & \textbf{62.92} & 69.31 & \textbf{75.54} & \textbf{75.15}
  & 52.53 & 50.00  & 44.62 & 49.60 \\
\Xhline{1.0pt}
\end{tabular}
}
\end{table*}
\renewcommand{\arraystretch}{1.0}

\subsubsection{Ablation Study}
We ablate each component of EgoMemo on EgoServe to understand its individual contribution. Results are reported in Table~\ref{tab:mainresult}.

Replacing the three-level temporal hierarchy with a single-scale caption store (\textit{w/o MS}) reduces the overall F1 from 8.0 to 7.0, with Memory Link dropping from 4.9 to 1.9 and Routine Optimization from 11.8 to 5.7, confirming that the hierarchical organization of clip-, activity-, and session-level summaries is essential for capturing patterns across extended temporal spans.
Removing the VLM-based caption reconstruction step (\textit{w/o Recons.}) causes a large overall drop (8.0 $\to$ 6.6). Both Memory Recall and Memory Link fall to 0.0, indicating that raw retrieved snippets are insufficient for the reasoner to synthesize coherent cross-temporal evidence. 
%
Disabling the visual embedding archive (\textit{w/o VSR}) yields a modest reduction (8.0 $\to$ 6.9), with the impact concentrated on Memory Link (4.9 $\to$ 3.2) and Routine Optimization (11.8 $\to$ 6.4), where visual cues help identify recurring objects or scenes that textual descriptions alone may miss.
Removing the knowledge graph pathway (\textit{w/o GSR}) reduces F1 to 6.5, with Memory Link dropping sharply to 0.0, highlighting the graph's role in linking semantically related entities across time.
Without multi-scale temporal retrieval (\textit{w/o MTR}), F1 drops to 6.8, and Memory Link falls to 2.1, as the system loses its primary mechanism for locating temporally relevant context at the appropriate granularity.

Taken together, caption reconstruction and graph semantic retrieval have the largest individual impact, while all three retrieval pathways (temporal, semantic, and visual) contribute complementary evidence. No single pathway subsumes another, validating the design of parallel heterogeneous retrieval for proactive assistance.

\subsubsection{Qualitative Analysis}
\begin{figure}
    \centering
    \includegraphics[width=0.9\linewidth]{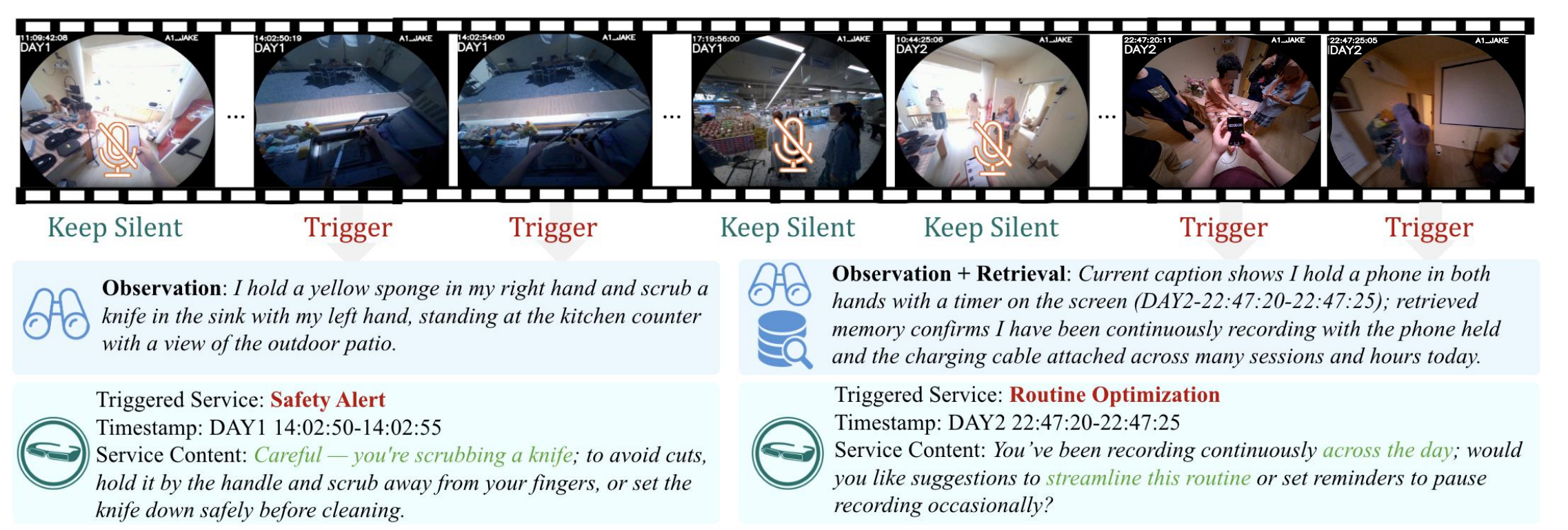}
    \caption{\textbf{Qualitative examples of EgoMemo's proactive assistance.} Left: an instant Safety Alert service triggered by observing the user scrubbing a knife barehanded. Right: a long-term Routine Optimization service triggered on Day 2 by detecting a recurring pattern of prolonged phone recording across multiple sessions and days.}
    \label{fig:qualitative}
\end{figure}

Fig.~\ref{fig:qualitative} shows two representative examples of EgoMemo's proactive behavior on EgoServe. In the left example, the agent observes the user scrubbing a knife and immediately triggers a Safety Alert based solely on the current clip-level caption. In the right example, the agent detects on Day~2 that the user has been holding a phone with a timer on screen; it retrieves cross-session memory confirming a recurring pattern of prolonged recording across multiple days, and triggers a Routine Optimization service suggesting ways to streamline this behavior. 

\subsection{Generalization to Existing Benchmarks}
We further evaluate EgoMemo on five established benchmarks to verify that its architecture generalizes beyond proactive assistance.

\renewcommand{\arraystretch}{1.2}
\begin{wraptable}{r}{0.46\textwidth}
\centering
\vspace{-8mm}
\caption{\small{\textbf{Results on offline egocentric video benchmarks.} For EgoTaskQA, we convert the dataset into MCQ format following~\cite{zhang2025exo2ego}.}}
\label{tab:offline benchmark}
\resizebox{0.47\textwidth}{!}{
\setlength{\tabcolsep}{6pt}
\begin{tabular}{l | c c c}
\Xhline{1.0pt}
\rowcolor{mygray}
\multirow{-1}{*}{\textbf{Method}}
  & \makecell{\textit{Ego-}\\\textit{TaskQA}} & \makecell{\textit{QA-}\\\textit{Ego4D}} & \makecell{\textit{Ego-}\\\textit{Schema}} \\
\hline
LLaVA-OneVision-7B~\cite{li2024llava}   & 55.8          & 65.7          & 60.1          \\
VideoLLaMA3~\cite{zhang2025videollama}        & 56.6          & 62.4          & 61.1          \\
VideoChat2-HD~\cite{li2024mvbench}      & 45.5          & 52.0          & 55.8          \\
Exo2Ego-7B~\cite{zhang2025exo2ego}        & 48.1          & 62.1          & 61.3          \\
Qwen2-VL-7B~\cite{bai2023qwen}               & 57.9          & 60.3          & 63.3          \\
EgoThinker~\cite{pei2025egothinker}                                & \textbf{64.4} & \underline{66.2} & \underline{67.6} \\
\Xhline{0.5pt}
\rowcolor{lightpurple}
\textbf{EgoMemo (Ours)}                   & \underline{60.9} & \textbf{68.0} & \textbf{74.8} \\
\Xhline{1.0pt}
\end{tabular}
}
\vspace{-7mm}
\end{wraptable}
\renewcommand{\arraystretch}{1}

\noindent\textbf{ESTP-Bench} (Table~\ref{tab:ESTP_result}). EgoMemo achieves the best overall score of 27.6 on explicit proactive tasks, surpassing EyeWO (23.6), with particularly strong gains on context-dependent subtasks such as TRU (32.4 vs.\ 25.1) and EOSC (35.8 vs.\ 20.8). On implicit proactive tasks, EgoMemo scores 34.7 compared to EyeWO's 52.5. Notably, EyeWO is the only trained model in this comparison; all others, including EgoMemo, are entirely training-free.

\noindent\textbf{OVO-Bench} (Table~\ref{tab:ovobench}). EgoMemo achieves the best real-time perception score of 75.15, outperforming GPT-4o (64.46) and LLaVA-OneVision (64.02). On backward tracing, EgoMemo scores 49.60, competitive with GPT-4o (60.75); the gap is expected given the substantially larger capacity and context windows of proprietary models.

\noindent\textbf{Offline egocentric QA} (Table~\ref{tab:offline benchmark}). EgoMemo achieves the best scores on two of the three egocentric QA datasets: EgoSchema (74.8, +7.2 over EgoThinker) and QAEgo4D (68.0 vs.\ 66.2), demonstrating that structured memory access is particularly effective for long-form temporal reasoning. On EgoTaskQA, EgoMemo scores 60.9, competitive with EgoThinker (64.4), where fine-grained action-state reasoning within short procedural segments favors direct visual perception over memory retrieval.

These results confirm that EgoMemo's streaming-first design generalizes across both streaming and offline settings without architectural modification.

\section{Conclusion}
We presented EgoServe, the first large-scale benchmark for proactive assistance in continuous egocentric video, comprising over 3,000 service instances across 10 categories and 4 temporal memory horizons. Alongside the benchmark, we introduced EgoMemo, a training-free, memory-augmented agent that maintains multi-scale temporal summaries, an evolving knowledge graph, and a visual embedding archive to perform retrieval-augmented reasoning over streaming video. Experiments show that EgoMemo establishes strong baselines on EgoServe and achieves competitive or state-of-the-art results on 5 existing benchmarks.

\noindent\textbf{Limitations and future work.}
EgoMemo's text-based memory loses fine-grained visual details during captioning, and the name-based entity resolution may fail for visually ambiguous entities. The semi-automated annotation pipeline may also introduce biases toward service types that are easier for foundation models to generate. Future directions include learning intervention timing from human preferences, incorporating audio context, and extending EgoServe to multi-turn proactive dialogues and multi-user settings.

\section{Acknowledgement}
The paper is supported in part by the National Natural Science Foundation of China under grant No.62441231, 62293542, Liao Ning Province Science and Technology Plan No.2023JH26/10200016, Dalian City Science and Technology Innovation Fund No.2023JJ11CG001, and Ningbo Key R\&D project under Grant No.2025Z039.

\section*{A. EgoServe Benchmark}

We first compare EgoServe with existing video understanding benchmarks in Table~\ref{tab:benchmark_comparison}. EgoServe is the first benchmark that simultaneously covers egocentric perspective, multi-day temporal span, proactive service evaluation, and streaming inference protocol, comprising 3.4k service instances across $\sim$128h of video from three complementary source datasets.

\renewcommand{\arraystretch}{1.2}
\begin{table}[h]
\centering
\caption{Comparison between EgoServe and existing video understanding benchmarks.}
\label{tab:benchmark_comparison}
\vspace{-3mm}
\small
\setlength{\tabcolsep}{3pt}
\resizebox{1.0\linewidth}{!}{
\begin{tabular}{l|cc|ccccc}
\Xhline{1.0pt}
\rowcolor{mygray}
\multicolumn{1}{c|}{\textbf{Benchmark}} & \textbf{\#Pairs} & \multicolumn{1}{c|}{\textbf{Total Len}} & \textbf{Ego} & \textbf{Multi-Day} & \textbf{Proactive} & \textbf{Streaming} \\
\hline
\multicolumn{1}{c|}{OVO-Bench~\cite{niu2025ovo}}  &  2.8k  & -  & \textcolor{red}{\ding{55}} & \textcolor{red}{\ding{55}} & \textcolor{red}{\ding{55}} & \textcolor{green!60!black}{\ding{51}} \\
\multicolumn{1}{c|}{StreamingBench~\cite{lin2024streamingbench}}  & 4.5k   & - & \textcolor{red}{\ding{55}} & \textcolor{red}{\ding{55}} & \textcolor{red}{\ding{55}} & \textcolor{green!60!black}{\ding{51}} \\
\multicolumn{1}{c|}{EgoSchema~\cite{mangalam2023egoschema}}  & 5k & $\sim$250h & \textcolor{green!60!black}{\ding{51}} & \textcolor{red}{\ding{55}} & \textcolor{red}{\ding{55}} & \textcolor{red}{\ding{55}} \\
\multicolumn{1}{c|}{EgoLife~\cite{yang2025egolife}}  & 3k & $\sim$266h & \textcolor{green!60!black}{\ding{51}} & \textcolor{green!60!black}{\ding{51}} & \textcolor{red}{\ding{55}} & \textcolor{red}{\ding{55}} \\
\multicolumn{1}{c|}{ProAssist~\cite{zhang2025proactive}}      & 30.1k & $\sim$478h  & \textcolor{green!60!black}{\ding{51}} & \textcolor{red}{\ding{55}} & \textcolor{green!60!black}{\ding{51}} & \textcolor{green!60!black}{\ding{51}} \\
\hline
\multicolumn{1}{c|}{\textbf{EgoServe (Ours)}}   & 3.4k & $\sim$128h & \textcolor{green!60!black}{\ding{51}} & \textcolor{green!60!black}{\ding{51}} & \textcolor{green!60!black}{\ding{51}} & \textcolor{green!60!black}{\ding{51}} \\
\Xhline{1.0pt}
\end{tabular}
}
\vspace{-5mm}
\end{table}
\renewcommand{\arraystretch}{1.2}

\subsection*{A.1 Annotation Details}
We describe the annotation procedure for each source dataset in detail.
As illustrated in Fig.~2 of the main paper, the annotation pipeline follows a semi-automated approach: category-specific prompts guide Gemini-2.5-Pro~\cite{team2023gemini} to generate candidate proactive service instances from existing human annotations, and all generated candidates undergo manual verification before inclusion in the final benchmark.
The complete prompt templates for all service categories are provided at the end of this supplementary material.

\subsubsection*{EgoLife Subset }
EgoLife provides multi-day continuous egocentric recordings with dense captions spanning the first five days of three participants (A1, A4, A5).
As shown in Fig.~2 (a) of the main paper, the annotation pipeline follows two distinct processing paths depending on the temporal horizon of the target service: the upper path handles Instant, Short-Term, and Episodic services, while the lower path handles Long-Term services.

\noindent\textbf{Stage 1: Preparing 1-hour human annotations.}
Initially, we segment each participant's daily recording into non-overlapping 1-hour intervals.
For each interval, we collect all available human annotations---including dense captions, action descriptions, speaker transcripts, and interaction logs---and organize them in chronological order as a structured JSON document.
This document, together with a summarization prompt, is streamed into the Gemini Annotator to produce a condensed activity record for the interval, capturing key events, objects, locations, and social interactions.
These interval-level summaries serve as the input to the second stage.

\noindent\textbf{Stage 2: Proactive service generation.}
For \textbf{Instant, Short-Term, and Episodic} services (the upper path in Fig.~2(a)), each 1-hour interval is processed independently.
The interval's structured annotations are streamed into the Gemini Annotator together with a service-specific prompt (e.g., Safety Alert, Error Recovery, Task Reminder).
The model directly triggers candidate service instances from the observations within that interval, producing structured JSON output conforming to a predefined schema that enforces fields such as temporal trigger window, service category, observation context, and proactive response.

For \textbf{Long-Term} services (Habit Coaching, Memory Link, Routine Optimization, Personal Feedback), we follow the lower path in Fig.~2 (a) and adopt the streaming cue-capturing strategy described in Sec.~3.4 of the main paper.
Rather than processing each interval independently, the Gemini Annotator first extracts cross-segment long-term events from each group of intervals, then carries the accumulated events to the next turn as additional context.
Concretely, for each participant, the pipeline proceeds as follows:
\begin{enumerate}
    \item The first group of intervals (e.g., Day~1) is processed with the category-specific prompt. The Gemini Annotator identifies potential long-term events (e.g., recurring behaviors, unfinished plans) and outputs both cross-segment long-term events (new observations relevant to the service type) and candidate triggers (proactive service instances where sufficient evidence has accumulated).
    \item The extracted cross-segment long-term events are carried to the next turn: they are prepended to the prompt for the next group of intervals. This allows the model to reason over cross-session patterns---for example, a Habit Coaching trigger on Day~3 may reference behavioral patterns first observed on Day~1.
    \item This process repeats until all days have been processed, progressively building a richer context for long-term service detection.
\end{enumerate}
This streaming design ensures that long-term services are grounded in genuine multi-day patterns rather than being fabricated from single-interval observations.

\subsubsection*{HoloAssist Subset}

HoloAssist captures task-oriented interactions where an instructor guides a user through procedural activities (e.g., assembling furniture, configuring devices).
The dataset provides structured human annotations including step boundaries, instructor corrections, and error flags.

Unlike EgoLife, HoloAssist annotations are processed in a single stage, as the task-oriented nature of the videos and the availability of fine-grained procedural annotations make direct service generation feasible.
For each of the selected videos, we preprocess the full set of human annotations in chronological order and construct a structured JSON document containing the complete interaction timeline---step descriptions, instructor utterances, error labels, and step transitions.

This document is then paired with a category-specific prompt and passed to Gemini.
Due to the task-oriented nature of HoloAssist, annotations primarily cover \textbf{Instant} and \textbf{Short-Term} service categories:
\begin{itemize}
    \item \textbf{Safety Alerts (SA):} triggered when the user handles potentially dangerous tools or materials in an unsafe manner.
    \item \textbf{Tool Use (TU):} triggered when the user could benefit from using a more appropriate tool or technique.
    \item \textbf{Next-Step Guidance (NSG):} generated from step transition points, where the model produces guidance for the upcoming procedural step.
    \item \textbf{Error Recovery (ER):} derived from instructor correction annotations, where the model generates corrective guidance when a procedural mistake is detected.
    \item \textbf{Resource Reminder (RR):} triggered when the user is about to leave a resource unattended or a closure task incomplete, such as leaving a door open or walking away from an unsaved document.
\end{itemize}
Each service category is annotated in a separate pass with its own prompt template and structured output schema, ensuring category-specific detection criteria and mutual exclusion rules are enforced.


\subsubsection{CaptainCook4D Subset}

CaptainCook4D provides structured cooking recordings with both correct and erroneous recipe executions across 24 recipes.
We select 87 videos from the validation and test splits that contain explicit step-error annotations.

The annotation process leverages the procedural step annotations and error labels to generate proactive service instances focused on three categories: \textbf{Tool Use (TU)}, \textbf{Next-Step Guidance (NSG)} and \textbf{Error Recovery (ER)}.
A single unified prompt instructs Gemini to map procedural step transitions to next-step guidance and error labels to corrective feedback.
The structured recipe context (correct step sequences and annotated deviations) provides strong supervision for generating temporally precise and factually grounded service instances.

\subsubsection*{Output Schema and Quality Control}
All annotation prompts enforce structured JSON output, where each service instance includes a temporal trigger window, service category label, observation context, and a proactive response. Long-Term service outputs additionally contain linked past events and cross-session reasoning chains. Category-level mutual exclusion rules are embedded in the prompts to prevent overlapping annotations, complementing the manual verification process described in Sec.~\ref{sec:supp_verification}.

We provide visualization examples of the generated annotations for all 11 service subcategories in Fig.~\ref{fig:instant_vis}--\ref{fig:long_term_vis}, organized by temporal horizon: Instant services (Safety Alert, Tool Use), Short-Term services (Error Recovery, Next-Step Guidance, Resource Reminder), Episodic services (Memory Recall, Task Reminder), and Long-Term services (Habit Coaching, Memory Link, Routine Optimization).

These visualizations illustrate two key properties of our annotations.
First, \textbf{contextual grounding}: every generated response is directly anchored in the observed video content. For Instant and Short-Term services, the responses reference specific objects, actions, or states visible in the current frames (e.g., a Safety Alert identifying an exposed hazard, or an Error Recovery correcting a procedural mistake observed in the clip). For Episodic and Long-Term services, the responses additionally draw on evidence from temporally distant past events---the ``Past Linked Messages'' and ``Trigger Reason'' fields in each visualization explicitly show what earlier observations support the current intervention and how they are connected across time gaps ranging from minutes to days.
Second, \textbf{response effectiveness}: the generated proactive dialogues are phrased as natural, actionable suggestions rather than generic warnings. Each response addresses a specific user need at the moment of intervention: Episodic services such as Memory Recall and Task Reminder help the user recover forgotten context or resume interrupted activities, while Long-Term services such as Habit Coaching and Routine Optimization synthesize multi-day behavioral patterns into concrete improvement suggestions. This combination of temporal grounding and natural phrasing ensures that the annotations reflect realistic proactive assistance scenarios.


\renewcommand{\arraystretch}{1.2}
\begin{table*}[t]
\centering
\caption{Manual verification statistics for EgoServe annotations across all service categories. MV represents Manual Verification while Acc. Rate refers to acceptance rate. }
\label{tab:verification_rate}
\small
\vspace{-3mm}
\setlength{\tabcolsep}{3pt}
\resizebox{1.0\textwidth}{!}{
\begin{tabular}{l | cc | ccc | cc | ccc | c }
\Xhline{1.0pt}
\rowcolor{mygray}
  & \multicolumn{2}{c|}{\textbf{Instant}}
  & \multicolumn{3}{c|}{\textbf{Short-term}}
  & \multicolumn{2}{c|}{\textbf{Episodic}}
  & \multicolumn{3}{c|}{\textbf{Long-term}}
  & \\
\rowcolor{mygray}
\multirow{-2}{*}{\textbf{Metrics}}
  & \textit{SA} & \textit{TU}
  & \textit{NSG} & \textit{ER} & \textit{RR}
  & \textit{MR} & \textit{TR}
  & \textit{HC} & \textit{ML}  & \textit{RO}
  & \multirow{-2}{*}{\textbf{Overall}} \\
\hline
Before MV
  & 195 & 549 & 1392 & 1236 & 182 & 67 & 112 & 76 & 80 & 149 &  4038 \\
After MV
  & 241 & 510 & 1214 & 925 & 127 & 63 & 88 & 61 & 80 & 128 & 3437 \\
Acc. Rate (\%) & 100 & 92.9 & 87.2 & 74.8 & 69.8 & 94.0 & 78.6 & 80.3 & 100.0 & 85.9 & 85.1   \\
\Xhline{1.0pt}
\end{tabular}
}
\vspace{-3mm}
\end{table*}
\renewcommand{\arraystretch}{1.0}

\subsection*{A.2 Manual Verification}
\label{sec:supp_verification}
As described in Sec.~3.4 of the main paper, all generated annotation candidates undergo a manual verification process before inclusion in the final EgoServe benchmark.

\noindent\textbf{Verification procedure.}
Two trained annotators jointly review each candidate annotation. For every service instance, both annotators watch the corresponding source video segment together and discuss whether the candidate should be accepted. A candidate is accepted only when both annotators reach consensus; otherwise it is discarded. The primary rejection criteria are:
\begin{itemize}
    \item \textbf{Lack of practical utility:} the generated response does not provide genuinely useful assistance to the user. For example, a service that repeats information the user has already obtained (e.g., ``Based on the dance video you selected earlier, it is one minute and fifty-two seconds'' when the user is already reading the duration on screen) is rejected, as it offers no additional value.
    \item \textbf{Temporal misalignment:} the proposed trigger time window does not correspond to the actual video content---the triggering context described in the annotation is not present in the frames at the specified timestamps.
    \item \textbf{Insufficient visual grounding:} the service content is generated purely from speech transcripts or dialogue context without incorporating visual observations. Since proactive assistance in egocentric video should be grounded in what the user \emph{sees and does}, annotations that rely solely on audio information are excluded.
\end{itemize}

\noindent\textbf{Verification statistics.}
Table~\ref{tab:verification_rate} reports the manual verification results for each service category. Out of 4{,}038 candidates generated by the annotation pipeline, 3{,}437 remain after manual verification, yielding an overall acceptance rate of 85.1\%.
Among Instant services, Safety Alert (SA) is a special case: its count increases after verification (195 $\rightarrow$ 241) rather than shrinking, because we found that Gemini had misclassified a number of genuine safety events into other categories and reassigned them back to SA during manual review; we therefore report its acceptance rate as 100\%. Tool Use (TU) likewise attains a high rate of 92.9\%, as both are triggered by immediately observable events that are straightforward to verify.
Short-Term services show progressively lower rates: Next-Step Guidance (NSG) reaches 87.2\%, whereas Error Recovery (ER: 74.8\%) and especially Resource Reminder (RR: 69.8\%, the lowest overall) are harder to confirm, reflecting the difficulty of precisely identifying procedural mistakes and recently-used resources from visual context alone.
Episodic services remain fairly reliable, with Memory Recall (MR) at 94.0\% and Task Reminder (TR) at 78.6\%, indicating that the temporal linking between past and current events is generally sound. Among Long-Term services, Memory Link (ML) is fully retained (100.0\%) and Routine Optimization (RO) reaches 85.9\%, while Habit Coaching (HC) has the lowest rate at 80.3\%, as cross-session behavioral patterns are more susceptible to generating responses with insufficient practical utility.
Overall, these results confirm that our semi-automated pipeline produces high-quality candidates, and that manual verification not only filters unreliable instances but also corrects category misassignments, ensuring the quality of the final EgoServe benchmark.





\section*{B. Methods}

\begin{figure*}
    \centering
    \includegraphics[width=1\linewidth]{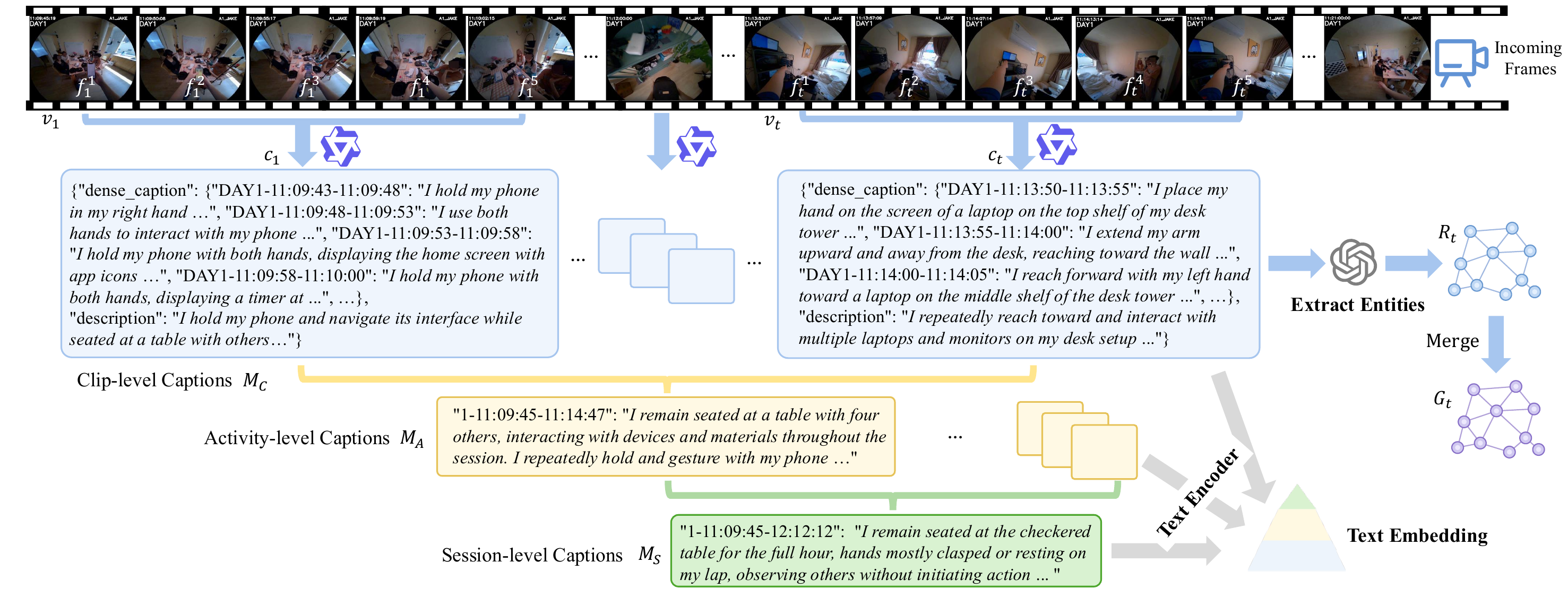}
    \vspace{-4mm}
    \caption{Illustration of the multi-scale temporal memory construction. Clip-level captions $\mathcal{M}_C$ preserve fine-grained timestamped action descriptions, which are progressively summarized into activity-level ($\mathcal{M}_A$) and session-level ($\mathcal{M}_S$) captions covering broader temporal spans.}
    \label{fig:memory_construction}
\end{figure*}

\subsection*{B.1 Details on Multi-Scale Temporal Memory Construction}

We provide additional details on the multi-scale temporal memory described in Sec.~4.1 of the main paper, with the concrete caption formats illustrated in Fig.~\ref{fig:memory_construction}.

\noindent\textbf{Clip-level Captions $\mathcal{M}_C$.}
For each incoming video clip $v_t$, a vision-language model produces a structured JSON object containing two fields:
(1)~\texttt{dense\_caption}, a dictionary mapping fine-grained timestamp intervals (e.g., "DAY1-11:09:43-11:09:48") to per-interval action descriptions, and
(2)~\texttt{description}, a brief summary that captures the overall activity of the clip.
Each timestamp follows the format \texttt{DAY\{d\}-HH:MM:SS-HH:MM:SS}, which jointly encodes the day index and the absolute start/end time, enabling precise temporal localization across multi-day recordings.
Fig.~\ref{fig:memory_construction} illustrates a concrete example from the EgoLife subset, where the clip-level captions effectively preserve fine-grained perceptual details such as object interactions and hand poses.

\noindent\textbf{Activity-level Captions $\mathcal{M}_A$.}
As the clip-level caption stream grows, we periodically aggregate consecutive clip captions within a temporal window $\mathcal{W}_j$ into an activity-level summary $M_A^{(j)}$ via an LLM-based summarize operation (Eq.~1 in the main paper).
Each activity-level caption is stored with a merged timestamp (e.g., "1-11:09:45-11:14:47") spanning the full window duration, and condenses the fine-grained clip-level details into a coherent activity narrative---for example, \emph{``I remain seated at a table with four others, interacting with devices and materials throughout the session.''}
This level captures medium-range behavioral context while significantly reducing memory size.

\noindent\textbf{Session-level Captions $\mathcal{M}_S$.}
Similarly, activity-level summaries are further aggregated into session-level summaries $M_S^{(k)}$ over a larger window $\mathcal{W}_k$.
Session-level captions cover a broader temporal span (e.g., "1-11:09:45-12:12:12" in the EgoLife subset) and distill the activity patterns into high-level descriptions of the user's overall routine and behavioral state.
This coarsest granularity allows the system to maintain a compact representation of long-horizon context spanning hours to days.

In addition to the temporal hierarchy, we extract named entities (people, objects, locations) from each clip caption $c_t$ to form the entity set $R_t$ (right side of Fig.~\ref{fig:memory_construction}), which are incrementally merged into the evolving knowledge graph $\mathcal{G}$ to support structure-aware semantic retrieval.
All captions at each level are encoded into dense embeddings $\{e^c_i, e^a_j, e^s_k\}$ using a text encoder \texttt{TEnc} and indexed for similarity search during retrieval.
The entire memory construction process---including caption generation, multi-scale summarization, entity extraction, and embedding indexing---operates in a fully streaming fashion: only newly accumulated segments trigger summarization at the next level, rather than re-processing the entire history, ensuring that both computational cost and latency remain bounded as the video stream extends over multiple days.



\subsection*{B.2 More Details on Different Benchmarks}
\label{sec:supp_benchmarks}


We provide detailed configurations for adapting EgoMemo to each evaluation benchmark, covering the proactive EgoServe benchmark, the online OVO-Bench, and all offline egocentric benchmarks.
As shown in Fig.~3 (b) of the main paper, at each reasoning step, the agent first decides whether retrieval is needed. This decision is made by the reasoning LLM itself through a structured prompt that instructs it to assess whether the available streaming context is sufficient to respond, or whether additional historical evidence must be retrieved. If retrieval is triggered, the agent generates a retrieval query and invokes the multi-pathway retrieval pipeline described in Sec.~4.2.

\subsubsection*{EgoServe (EgoLife Subset)}
The EgoLife subset of EgoServe consists of multi-day, multi-hour egocentric recordings. The memory construction and retrieval process follows the pipeline described in Sec.~4 of the main paper almost exactly, with all three levels of temporal memory ($\mathcal{M}_C$, $\mathcal{M}_A$, $\mathcal{M}_S$) constructed and maintained in a streaming fashion.

At each reasoning step, the streaming input to the reasoning agent consists of the current clip-level caption $c_t$ together with the most recent activity-level caption $M_A^{(j)}$, which provides broader situational context (e.g., what the user has been doing over the past several minutes) beyond the fine-grained details of the current clip alone. This combination allows the reasoner to make more informed intervention decisions and to generate temporally scoped retrieval queries when additional historical context is needed.

Specifically, for \textbf{time-aware retrieval query generation}: when the reasoning agent determines that retrieval is necessary, it is additionally prompted to include a temporal scope descriptor (e.g., ``last one hour'', ``last day'') within the query. A separate LLM call then resolves this relative temporal reference against the known start and end times of each day's recording, producing an absolute time range (e.g., DAY1-11:30:00 to DAY3-14:15:00). The subsequent retrieval is then restricted to captions falling within this resolved time window, which substantially reduces the search space for long multi-day recordings while preserving the ability to recall distant episodic events.


\subsubsection*{OVO-Bench}

OVO-Bench evaluates online video understanding across two main question categories: realtime perception and backward tracing.

\noindent\textbf{Realtime perception.}
For realtime questions, the model must answer based on what is currently happening in the video. We directly provide the \textbf{clip-level captions from the last 10 seconds} together with \textbf{all available activity-level and session-level summaries} to the reasoning agent, which outputs the answer without any retrieval step. This design reflects the nature of realtime queries: the answer lies in the most recent observations, while the global summaries supply sufficient contextual grounding.

\noindent\textbf{Backward tracing.}
For backward tracing questions, the model must reason about events that occurred earlier in the video. We adopt a coarse-to-fine inference strategy: all activity-level and session-level summaries are provided as global context, and the agent first attempts to answer from this high-level information. If the agent judges the context insufficient, it generates a retrieval query and invokes the retrieval pipeline to fetch fine-grained clip-level evidence, following the same iterative retrieve-and-reason process.

\begin{table*}[t]
\centering
\caption{Detailed evaluation results of Precision (P) and Recall (R) on EgoServe benchmark. \textbf{w/o MS} replaces the three-level temporal memory hierarchy with a single clip-level caption store while \textbf{w/o Recons.} removes the VLM-based caption reconstruction step. }
\label{tab:egoserve_pr}
\vspace{-3mm}
\small
\setlength{\tabcolsep}{3pt}
\resizebox{1.0\textwidth}{!}{
\begin{tabular}{l | cccc | cccccc | cccc | cccccc | cc }
\Xhline{1.0pt}
\rowcolor{mygray}
  & \multicolumn{2}{c}{\textit{SA}} & \multicolumn{2}{c|}{\textit{TU}}
  & \multicolumn{2}{c}{\textit{NSG}} & \multicolumn{2}{c}{\textit{ER}} & \multicolumn{2}{c|}{\textit{RR}}
  & \multicolumn{2}{c}{\textit{MR}} & \multicolumn{2}{c|}{\textit{TR}}
  & \multicolumn{2}{c}{\textit{HC}} & \multicolumn{2}{c}{\textit{ML}}  & \multicolumn{2}{c|}{\textit{RO}}
  & \multicolumn{2}{c}{\textbf{AVG}} \\
\rowcolor{mygray}
\multirow{-2}{*}{\textbf{Model}} & P & R & P & R  & P & R & P & R & P & R  & P & R & P & R & P & R & P & R & P & R  & P & R \\

\hline
Qwen3-VL-Plus & \textbf{8.8} & 3.7 & 12.9 & 0.8 & 9.0 & 8.2 & 17.2 & \textbf{7.5} & 1.7 & \textbf{86.6} & 0.0 & 0.0 & 3.9 & 1.1 & \textbf{6.7} & 3.3 & 0.0 & 0.0 & 0.0 & 0.0 & 6.0 & 11.1 \\
GPT-5-mini & 7.0 & \textbf{59.8} & 7.7 & 2.4 & 8.5 & 10.9 & 10.6 & 0.5 & 2.7 & 59.8 & 0.0 & 0.0 & \textbf{11.5} & 8.0 & 5.0 & 6.6 & 0.0 & 0.0 & 0.0 & 0.0 & 5.3 & 14.8 \\
\hline
w/o MS & 6.3 & 46.9 & 18.3 & 5.1 & 18.3 & 35.8 & 21.4 & 1.0 & 2.6 & 35.4 & 3.8 & \textbf{6.3} & 1.8 & 6.8 & 2.8 & \textbf{13.1} & 3.7 & 1.2 & 4.4 & 7.8 & 8.3 & 15.9 \\
w/o Recons. & 6.0 & 43.1 & 20.2 & 4.7 & \textbf{19.9} & \textbf{38.0} & 6.9 & 0.2 & \textbf{3.1} & 40.9 & 0.0 & 0.0 & 3.3 & 9.1 & 3.8 & 9.8 & 0.0 & 0.0 & 5.3 & 6.2 & 6.8 & 15.2 \\
w/o VSR & 6.7 & 45.2 & 15.1 & 3.7 & 18.9 & 36.2 & 14.3 & 0.7 & 2.1 & 27.6 & 1.0 & 1.6 & 4.1 & \textbf{17.1} & 2.5 & \textbf{13.1} & 4.4 & 2.5 & 4.6 & 10.2 & 7.4 & 15.8 \\
w/o GSR & 7.2 & 49.0 & 14.0 & 3.5 & 18.4 & 36.3 & \textbf{30.0} & 1.0 & 2.6 & 36.2 & 2.3 & 3.2 & 3.0 & 11.4 & 1.5 & 6.6 & 0.0 & 0.0 & 4.1 & 7.8 & 8.3 & 15.5 \\
w/o MTR & 5.5 & 40.3 & \textbf{21.9} & \textbf{5.5} & 18.6 & 36.2 & 15.0 & 0.7 & 2.2 & 29.1 & \textbf{5.4} & 3.2 & 3.1 & 8.0 & 1.4 & 3.3 & 5.9 & 1.3 & 7.3 & 7.8 & 8.6 & 13.5 \\
\hline
\rowcolor{lightpurple}
\textbf{EgoMemo (Ours)} & 6.5 & 48.6 & 19.1 & 4.7 & 18.7 & 36.3 & 22.2 & 0.9 & 2.5 & 33.9 & 3.1 & 4.8 & 3.5 & 14.8 & 2.2 & 11.5 & \textbf{7.1} & \textbf{3.8} & \textbf{8.7} & \textbf{18.8} & \textbf{9.4} & \textbf{17.8} \\

\Xhline{1.0pt}
\end{tabular}
}
\vspace{-3mm}
\end{table*}

\subsubsection*{Offline Egocentric Benchmarks}

We provide the detailed configuration for each offline benchmark as follows.

\noindent\textbf{EgoSchema. }
Each video in EgoSchema is approximately 3 minutes long. Given this moderate duration, we construct only two levels of temporal memory: \textbf{clip-level} captions $\mathcal{M}_C$ with a window size of 15 seconds, and \textbf{activity-level} summaries $\mathcal{M}_A$ with a window size of 1 minute, yielding roughly 12 clip-level captions and 3 activity-level summaries per video. Session-level summaries $\mathcal{M}_S$ are omitted as the video length does not warrant a third level of abstraction.

During the caption generation phase, we provide the VLM with not only the sampled video frames of each clip but also the question and its candidate answer options. This encourages the VLM to attend to visual details that are relevant to the downstream question, producing more informative and targeted clip-level descriptions than generic captions would offer.

At inference time, the agent first receives all three activity-level summaries as global context and attempts to answer the question based on this high-level information alone. If the agent determines that the available context is insufficient, it generates a retrieval query and invokes the retrieval pipeline (Sec.~4.2) to retrieve fine-grained clip-level evidence. This retrieve-and-reason cycle is repeated for up to 3 rounds, allowing the agent to progressively gather more specific evidence until a confident answer is reached.

\noindent\textbf{EgoTaskQA and QAEgo4D. }
EgoTaskQA and QAEgo4D are short-duration egocentric video understanding benchmarks, where video lengths range from a few seconds to several minutes. We construct temporal memory using a \textbf{clip-level} window of 10 seconds and an \textbf{activity-level} window of 1 minute, with session-level summaries again omitted due to the limited video duration. The question-aware captioning strategy described above is applied identically.

Since many videos in these benchmarks are very short (under 10 seconds), a fixed coarse-to-fine strategy would be unnecessarily complex. We therefore adopt an adaptive approach: the agent first checks whether any activity-level summaries exist for the given video. If no activity-level summary is available---indicating that the video is too short for even a single activity-level aggregation window---all clip-level captions are directly provided to the reasoning agent, which is prompted to output an answer without invoking retrieval. If activity-level summaries do exist, the inference procedure follows the same coarse-to-fine retrieval process as EgoSchema: the agent first attempts to answer from activity-level context, and resorts to iterative retrieval over clip-level captions when needed.

\section*{C. Experiments}

\renewcommand{\arraystretch}{1.2}
\begin{table*}[t]
\centering
\caption{Evaluation results on EgoLife subset.}
\label{tab:egolife_subset}
\small
\vspace{-3mm}
\setlength{\tabcolsep}{3pt}
\resizebox{1.0\textwidth}{!}{
\begin{tabular}{l | cc | ccc | cc | ccc | c }
\Xhline{1.0pt}
\rowcolor{mygray}
  & \multicolumn{2}{c|}{\textbf{Instant}}
  & \multicolumn{3}{c|}{\textbf{Short-term}}
  & \multicolumn{2}{c|}{\textbf{Episodic}}
  & \multicolumn{3}{c|}{\textbf{Long-term}}
  & \\
\rowcolor{mygray}
\multirow{-2}{*}{\textbf{Model}}
  & \textit{SA} & \textit{TU}
  & \textit{NSG} & \textit{ER} & \textit{RR}
  & \textit{MR} & \textit{TR}
  & \textit{HC} & \textit{ML} & \textit{RO}
  & \multirow{-2}{*}{\textbf{Overall}} \\
\hline
Qwen3-VL-Plus & 6.2 & 5.0 & 9.4 & \textbf{4.4} & 3.7 & 0.0 & 1.8 & 4.4 & 0.0 & 0.0 & 3.5 \\
GPT-5-mini & 10.3 & 0.0 & 8.6 & 0.0 & 6.7 & 0.0 & \textbf{9.4} & \textbf{5.7} & 0.0 & 0.0 & 4.1 \\
\hline
w/o MS & 14.3 & 8.0 & 7.6 & 0.0 & 9.6 & \textbf{4.8} & 2.9 & 4.6 & 1.9 & 5.7 & 5.9 \\
w/o Recons. & 13.0 & \textbf{9.1} & 9.5 & 0.0 & \textbf{11.9} & 0.0 & 4.8 & 5.4 & 0.0 & 5.7 & 5.9 \\
w/o VSR & 14.9 & 4.9 & 5.7 & 0.0 & 7.5 & 1.2 & 6.6 & 4.2 & 3.2 & 6.4 & 5.5 \\
w/o GSR & \textbf{18.5} & 0.0 & 7.3 & 0.0 & 9.7 & 2.6 & 4.8 & 2.4 & 0.0 & 5.4 & 5.1 \\
w/o MTR & 12.1 & 0.0 & \textbf{12.3} & 0.0 & 7.5 & 4.0 & 4.5 & 2.0 & 2.1 & 7.6 & 5.2 \\
\Xhline{0.5pt}
\rowcolor{lightpurple}
\textbf{EgoMemo (Ours)} & 18.2 & 4.4 & 8.8 & 0.0 & 8.8 & 3.8 & 5.7 & 3.7 & \textbf{4.9} & \textbf{11.9} & \textbf{7.0} \\

\Xhline{1.0pt}
\end{tabular}
}
\vspace{-3mm}
\end{table*}
\renewcommand{\arraystretch}{1.0}

\subsection*{C.1 Detailed Results on EgoServe Benchmark}

\noindent\textbf{Precision and Recall breakdown (Table~\ref{tab:egoserve_pr}).}
Table~\ref{tab:egoserve_pr} reports per-category Precision (P) and Recall (R) on the full EgoServe benchmark. Across all models, average Recall consistently exceeds Precision, indicating that the primary bottleneck lies in generating precisely timed and relevant interventions rather than missing service opportunities. Our full model (EgoMemo) achieves the highest average Precision (9.4) and Recall (17.8), confirming that the complete pipeline strikes a reasonable balance between proactive coverage and intervention quality. Among the baselines, GPT-5-mini exhibits a strong Recall bias (14.8) with very low Precision (5.3), suggesting frequent but poorly targeted interventions. 
\renewcommand{\arraystretch}{1.2}
\begin{wraptable}{r}{0.47\textwidth}
\centering
 \vspace{-8mm}
\caption{Evaluation results on Captaincook4d subset.}
\label{tab:captaincook4d_subset}
\small
\setlength{\tabcolsep}{5pt}
\resizebox{0.47\textwidth}{!}{
\begin{tabular}{l | cc | ccc | c }
\Xhline{1.0pt}
\rowcolor{mygray}
  & \multicolumn{2}{c|}{\textbf{Instant}}
  & \multicolumn{3}{c|}{\textbf{Short-term}}
  & \\
\rowcolor{mygray}
\multirow{-2}{*}{\textbf{Model}}
  & \textit{SA} & \textit{TU}
  & \textit{NSG} & \textit{ER} & \textit{RR}
  & \multirow{-2}{*}{\textbf{Overall}} \\
\hline
Qwen3-VL-Plus & 3.9 & 2.4 & \textbf{12.2} & \textbf{10.0} & \textbf{2.3} & 6.2 \\
GPT-5-mini & \textbf{22.8} & 8.3 & 7.7 & 1.5 & 2.0 & \textbf{8.4} \\
\hline
w/o MS & 12.9 & 9.9 & 9.5 & 2.0 & 1.8 & 7.2 \\
w/o Recons. & 12.1 & 11.3 & 9.6 & 0.5 & 1.8 & 7.1 \\
w/o VSR & 12.5 & 8.8 & 9.4 & 1.5 & 1.6 & 6.8 \\
w/o GSR & 13.7 & 8.5 & 10.3 & 0.5 & 2.0 & 7.0 \\
w/o MTR & 11.5 & \textbf{14.5} & 10.0 & 0.5 & 2.1 & 7.7 \\
\Xhline{0.5pt}
\rowcolor{lightpurple}
\textbf{EgoMemo (Ours)} & 12.2 & 11.0 & 9.3 & 1.5 & 2.1 & 7.2 \\
\Xhline{1.0pt}
\end{tabular}
}
\vspace{-8mm}
\end{wraptable}
\renewcommand{\arraystretch}{1.0}
Comparing ablation variants, removing the multi-scale temporal retrieval (w/o MTR) causes the largest Recall drop in Long-Term categories (e.g., RO: 7.8 vs.\ 18.8), while removing caption reconstruction (w/o Recons.) yields the largest drop in average Precision (6.8 vs.\ 9.4) and collapses Memory Recall and Memory Link to zero (MR: 0.0 vs.\ 4.8; ML: 0.0 vs.\ 3.8 in Recall), reflecting the importance of caption reconstruction for grounding episodic and long-term retrieval.

\noindent\textbf{Per-subset results (Tables~\ref{tab:egolife_subset}--\ref{tab:holoassist_subset}).}
Tables~\ref{tab:egolife_subset}--\ref{tab:holoassist_subset} break down the F1 scores across the three source datasets. 
On the EgoLife subset (Table~\ref{tab:egolife_subset}), 
EgoMemo achieves the best overall score (7.0), with notable advantages in Routine Optimization (11.9) and Memory Link (4.9); the multi-scale temporal retrieval (w/o MTR) and graph-based semantic retrieval (w/o GSR) ablations show the largest 
\renewcommand{\arraystretch}{1.2}
\begin{wraptable}{r}{0.47\textwidth}
\centering
\vspace{-10mm}
\caption{Evaluation results on HoloAssist subset.}
\label{tab:holoassist_subset}
\small
\setlength{\tabcolsep}{3pt}
\resizebox{0.47\textwidth}{!}{
\begin{tabular}{l | cc | ccc | cc | cccc | c }
\Xhline{1.0pt}
\rowcolor{mygray}
  & \multicolumn{2}{c|}{\textbf{Instant}}
  & \multicolumn{3}{c|}{\textbf{Short-term}}
  & \\
\rowcolor{mygray}
\multirow{-2}{*}{\textbf{Model}}
  & \textit{SA} & \textit{TU}
  & \textit{NSG} & \textit{ER} & \textit{RR}
  & \multirow{-2}{*}{\textbf{Overall}} \\
\hline

Qwen3-VL-Plus & 0.0 & 0.0 & 7.1 & \textbf{11.7} & 0.0 & 3.8 \\
GPT-5-mini & 0.0 & 0.0 & 11.0 & 0.8 & \textbf{0.6} & 2.5 \\
\hline
w/o MS & 1.1 & \textbf{6.0} & 36.1 & 1.9 & 0.0 & \textbf{9.0} \\
w/o Recons. & 1.2 & 3.5 & \textbf{38.0} & 0.4 & 0.0 & 8.6 \\
w/o VSR & \textbf{3.2} & 3.3 & 37.0 & 1.2 & 0.0 & 8.9 \\
w/o GSR & 2.2 & 3.4 & 36.2 & 3.2 & 0.0 & \textbf{9.0} \\
w/o MTR & 1.6 & 4.2 & 35.5 & 1.9 & 0.0 & 8.6 \\
\Xhline{0.5pt}
\rowcolor{lightpurple}
\textbf{EgoMemo (Ours)} & 1.2 & 4.7 & 36.3 & 2.0 & 0.0 & 8.8 \\

\Xhline{1.0pt}
\end{tabular}
}
\vspace{-8mm}
\end{wraptable}
\renewcommand{\arraystretch}{1.0}
drops, underscoring the importance of hierarchical context and entity semantics for 
multi-day recordings. On CaptainCook4D (Table~\ref{tab:captaincook4d_subset}), the scores are generally higher due to the structured procedural nature of cooking tasks,
with EgoMemo reaching 7.2 overall. 
Here GPT-5-mini yields the best overall score (8.4), demonstrating its strength in detecting safety-related events in cooking scenarios. 
On HoloAssist (Table~\ref{tab:holoassist_subset}), Next-Step Guidance dominates all models (36.3 for ours), while Instant and other Short-Term categories remain challenging. Notably, w/o GSR and w/o MS achieves the highest overall score 
(9.0) on this subset, suggesting that graph-based semantic retrieval and high-level captions provide limited benefit for short, self-contained procedural videos where the relevant context is mostly local.

\subsection*{C.2 Retrieval Time Analysis}
\label{sec:supp_inference_time}

To compare retrieval efficiency across methods, we randomly sample 30 videos from the EgoSchema test set and measure the average time (in seconds) each method requires to process one minute of video under a single-retrieval setting, 
where the model decides whether retrieval is needed and performs at most one retrieval pass.
As shown in Table~\ref{tab:retrieval_time}, VideoAgent~\cite{wang2024videoagent} requires 67.25 seconds per minute of video due to its iterative multi-round agent reasoning process.
The base Qwen2.5VL-7B~\cite{bai2025qwen25vltechnicalreport} model, which processes video frames directly without retrieval, achieves the lowest latency at 2.95 seconds, but lacks the ability to access historical context.
Augmenting it with Video-RAG~\cite{luo2024video} increases the retrieval time to 20.81 seconds owing to the additional retrieval and context integration overhead.
EgoMemo achieves a retrieval time of 13.11 seconds per minute of video, which is $5.1\times$ faster than VideoAgent and $1.6\times$ faster than Video-RAG, while supporting richer retrieval pathways (temporal, semantic, and visual).
\renewcommand{\arraystretch}{1.2}
\begin{wraptable}{r}{0.43\textwidth}
\centering
\vspace{-11mm}
\caption{Retrieval time comparison across different methods, measured in seconds per minute of video.}
\label{tab:retrieval_time}
\small
\setlength{\tabcolsep}{3pt}
\resizebox{0.43\textwidth}{!}{
\begin{tabular}{l | c }
\Xhline{1.0pt}
\rowcolor{mygray}
  \multicolumn{1}{c|}{\textbf{Method}}
  & \textbf{Retrieval Time (Sec)}
  \\
\hline
VideoAgent
  & 67.25  \\
Qwen2.5VL-7B
  & 2.95  \\
Qwen2.5VL-7B + Video-RAG
  & 20.81  \\
\Xhline{0.5pt}
\rowcolor{lightpurple}
\textbf{EgoMemo (Ours)}
  & 13.11 \\
\Xhline{1.0pt}
\end{tabular}
}
\vspace{-8mm}
\end{wraptable}
\renewcommand{\arraystretch}{1.0}
We note that EgoMemo's memory construction phase takes 77.33 seconds per minute of video on average; however, since memory construction and downstream reasoning operate asynchronously in our streaming pipeline, this cost does not add to the per-query retrieval latency.

\renewcommand{\arraystretch}{1.2}
\begin{table*}[t]
\centering
\caption{Detailed results of LLM evaluation on EgoServe benchmark. R represents the rationality and E refers to effectiveness. }
\label{tab:llm_evaluation}
\small
\vspace{-3mm}
\resizebox{1.0\textwidth}{!}{
\begin{tabular}{l|c|cccccccc}
\Xhline{1.0pt}
\rowcolor{mygray}
Eval LLM & \multicolumn{1}{c}{Metrics} & Qwen3-VL-Plus & GPT-5-mini & w/o MS & w/o Recons & w/o VSR & w/o GSR & w/o MTR & EgoMemo \\
\hline
\multirow{3}{*}{GPT-4o~\cite{gpt4o}} & 
 R & 2.5 & \textbf{2.7} & 2.5 & 2.5 & 2.5 & 2.5 & 2.5 & 2.5 \\
 & E & 3.0 & \textbf{3.3} & 3.0 & 3.0 & 3.0 & 3.0 & 2.9 & 3.0 \\
 & Overall & 2.8 & \textbf{3.0} & 2.8 & 2.8 & 2.8 & 2.8 & 2.7 & 2.8 \\
 \hline
\multirow{3}{*}{Deepseek-R1~\cite{guo2025deepseek}} & 
R & 2.3 & \textbf{2.5} & 2.2 & 2.2 & 2.2 & 2.2 & 2.2 & 2.2 \\
 & E & 2.5 & \textbf{2.9} & 2.6 & 2.4 & 2.4 & 2.5 & 2.3 & 2.5 \\
 & Overall & 2.4 & \textbf{2.7} & 2.4 & 2.3 & 2.3 & 2.4 & 2.2 & 2.3 \\
\Xhline{1.0pt}
\end{tabular}
}
\vspace{-3mm}
\end{table*}
\renewcommand{\arraystretch}{1.0}

\subsection*{C.3 Evaluation Protocol}
\label{sec:eval_protocol}
\noindent\textbf{Temporal matching.}
Each predicted service event is characterized by a trigger time window $(t^\text{start}_p, t^\text{end}_p)$ and a service sub-type $s_p$; ground-truth (GT) events are similarly defined as $(t^\text{start}_g, t^\text{end}_g, s_g)$.
To handle inherent temporal ambiguity in proactive service triggers, we introduce a dataset-specific tolerance $\delta$ that reflects the annotation granularity and typical action duration of each dataset: $\delta = 60$\,s for EgoLife (long-form daily activities with coarse temporal boundaries), $\delta = 10$\,s for HoloAssist (short procedural tasks), and $\delta = 25$\,s for CaptainCook4D (medium-length cooking sessions).
Matching is performed \textbf{per sub-type in isolation}: for each service sub-type $s$, we collect all predictions and GT events of type $s$ and apply greedy nearest-neighbor matching.
Concretely, for each GT event $g$, we compute the temporal center of every unmatched prediction $p$ and check whether it falls within the expanded window $[t^\text{start}_g - \delta,\; t^\text{end}_g + \delta]$.
Among all qualifying candidates, the prediction with the smallest distance to the GT interval is selected as the match, where the distance is zero if the prediction center lies within $[t^\text{start}_g, t^\text{end}_g]$ and equals the gap to the nearest boundary otherwise.
Each prediction can be matched to at most one GT event, enforcing a one-to-one assignment.

\begin{figure*}
    \centering
    \includegraphics[width=1\linewidth]{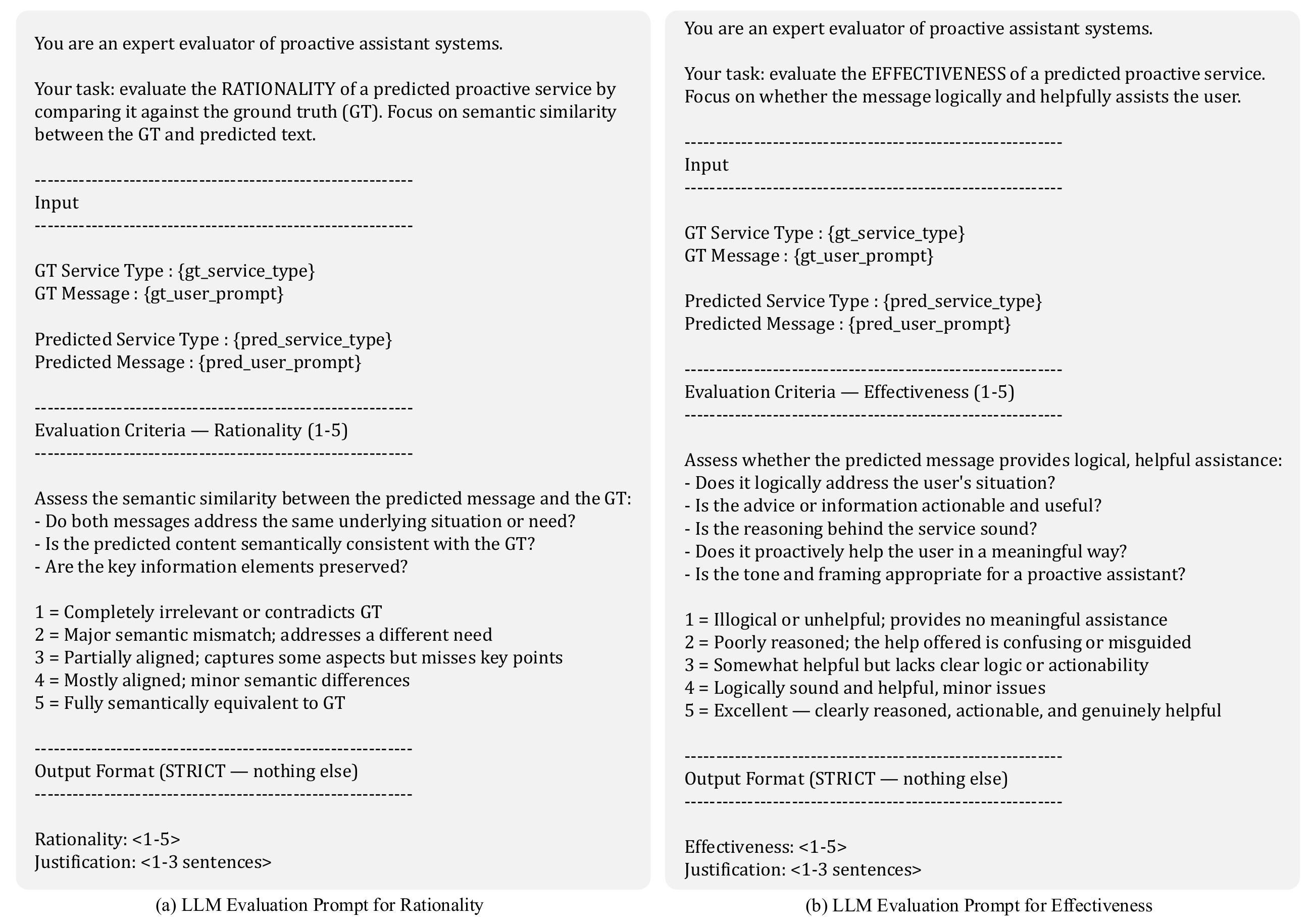}
    \vspace{-4mm}
    \caption{LLM Evaluation Prompts.}
    \vspace{-4mm}
    \label{fig:eval_prompts}
\end{figure*}

\noindent\textbf{Detection metrics.}
For each service sub-type $s$, we aggregate matched, predicted, and GT counts \emph{globally across all videos} in the dataset and compute Precision, Recall, and F1:
\begin{equation}
  \text{P}_s = \frac{|\text{matched}_s|}{|\text{pred}_s|},\quad
  \text{R}_s = \frac{|\text{matched}_s|}{|\text{gt}_s|},\quad
  \text{F1}_s = \frac{2\,\text{P}_s\,\text{R}_s}{\text{P}_s + \text{R}_s}.
\end{equation}
We report the \textbf{macro-averaged F1} over all active sub-types (those with at least one GT or prediction) as the primary metric, which prevents dominant sub-types from overshadowing rare ones. 

\noindent\textbf{LLM-as-judge scoring.}
Beyond detection accuracy, we evaluate the quality of generated service dialogues for successfully matched prediction--GT pairs.
Following recent practices in open-ended generation evaluation~\cite{zheng2023judging}, we employ an LLM as an automatic judge to assess two complementary dimensions on a 1--5 scale, as shown in Fig.~\ref{fig:eval_prompts}:
(1)~\textbf{Rationality} measures the semantic alignment between the predicted dialogue and the GT, considering whether both address the same underlying situation and preserve key informational elements;
(2)~\textbf{Effectiveness} evaluates whether the predicted dialogue provides logically sound, actionable, and genuinely helpful proactive assistance.
Each dimension is scored independently with temperature~0 to ensure reproducibility, and we report the average score across all matched pairs.



To further verify the robustness of our evaluation, we employ two independent LLM judges---GPT-4o~\cite{gpt4o} and DeepSeek-R1~\cite{guo2025deepseek}---and report the detailed Rationality (R) and Effectiveness (E) scores in Table~\ref{tab:llm_evaluation}. Although DeepSeek-R1 is systematically stricter, assigning lower absolute scores than GPT-4o (mean overall 2.38 vs.\ 2.81), the two judges produce highly consistent \emph{relative} rankings across all models and ablation variants: both identify the same best-performing configuration (the GPT-5-mini baseline) and the same weakest one (w/o MTR). The per-model overall scores of the two judges are strongly correlated (Pearson r $\approx$ 0.95), confirming that our conclusions are not sensitive to the choice of judge model. We further observe that, under both judges, Effectiveness scores are consistently higher than Rationality scores for all models, suggesting that the generated responses are generally helpful in tone and framing even when they do not precisely match the ground-truth semantics.

\subsection*{C.3 More Qualitative Results}
\label{sec:supp_qualitative}

\begin{figure}
    \centering
    \includegraphics[width=1\linewidth]{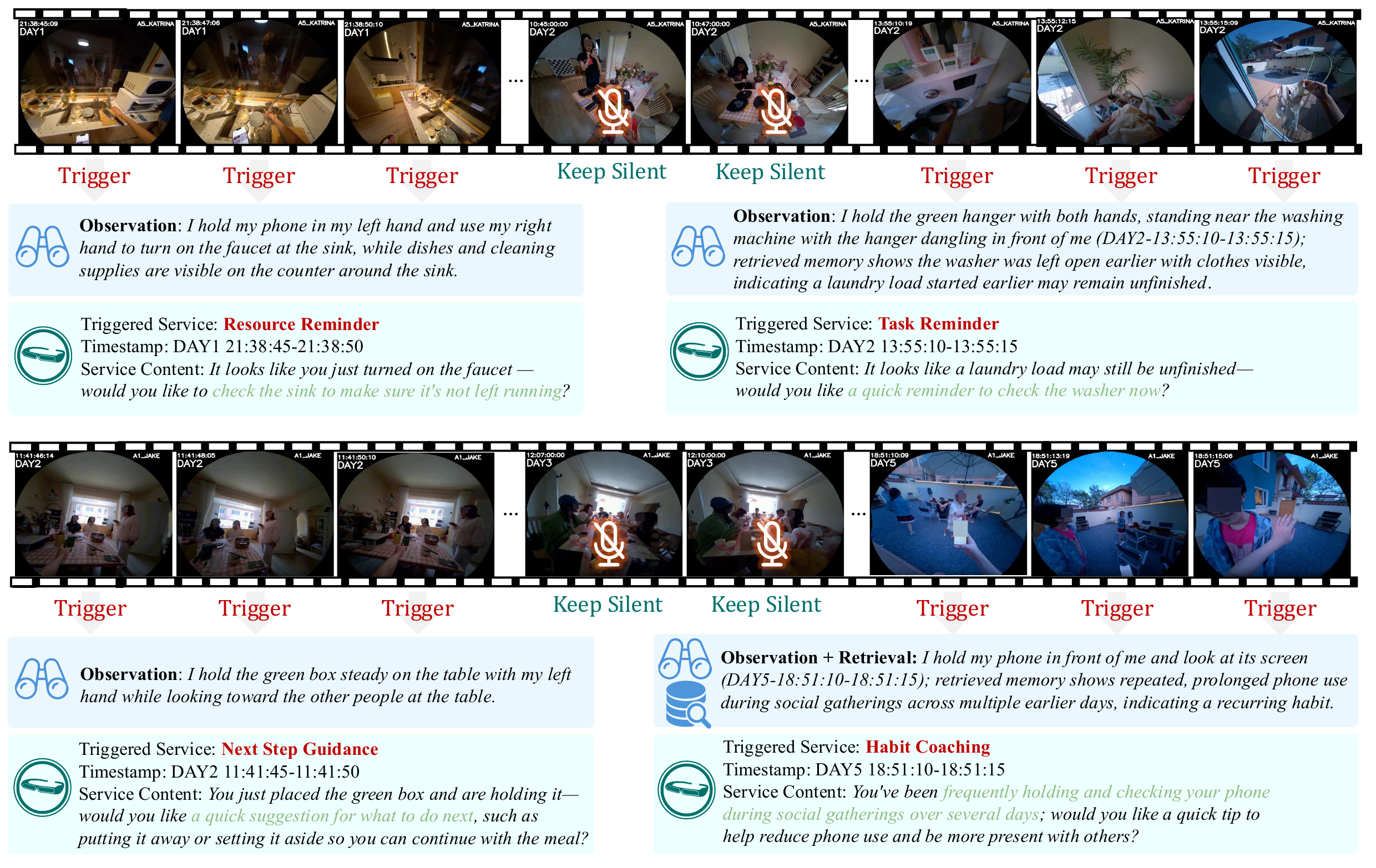}
    \vspace{-4mm}
    \caption{Qualitative examples of EgoMemo’s proactive assistance. }
    \vspace{-3mm}
    \label{fig:egomemo_vis}
\end{figure}

Fig.~\ref{fig:egomemo_vis} presents qualitative examples of EgoMemo's proactive assistance across four representative service categories on the EgoLife subset. Each example illustrates both the streaming decision process (Trigger vs.\ Keep Silent) and the generated service content.

The top-left example shows a \textbf{Resource Reminder} triggered on Day~1: the system observes the user turning on the faucet at the sink and proactively reminds them to check that the water is not left running before walking away. The top-right example demonstrates a \textbf{Task Reminder} on Day~4: after the user moves away from a table with unfinished flower-arranging materials, the system reminds them to complete or tidy up the activity. Both cases rely solely on recent clip-level observations to detect closure failures or interrupted tasks within a short temporal window.

The bottom two examples highlight services that require broader context. The \textbf{Next-Step Guidance} (bottom-left) is triggered on Day~2 when the user places a green box on the table during a meal, and the system suggests a natural follow-up action. The \textbf{Habit Coaching} example (bottom-right) demonstrates the value of long-term memory retrieval: on Day~5, the system detects the user checking their phone and retrieves evidence of repeated phone use during social gatherings across multiple earlier days, synthesizing this cross-day pattern into a coaching suggestion to reduce screen time. Notably, the streaming timeline shows that EgoMemo appropriately remains silent during intervals where no actionable service opportunity is detected, avoiding unnecessary interruptions.

\begin{figure}
    \centering
    \includegraphics[width=1\linewidth]{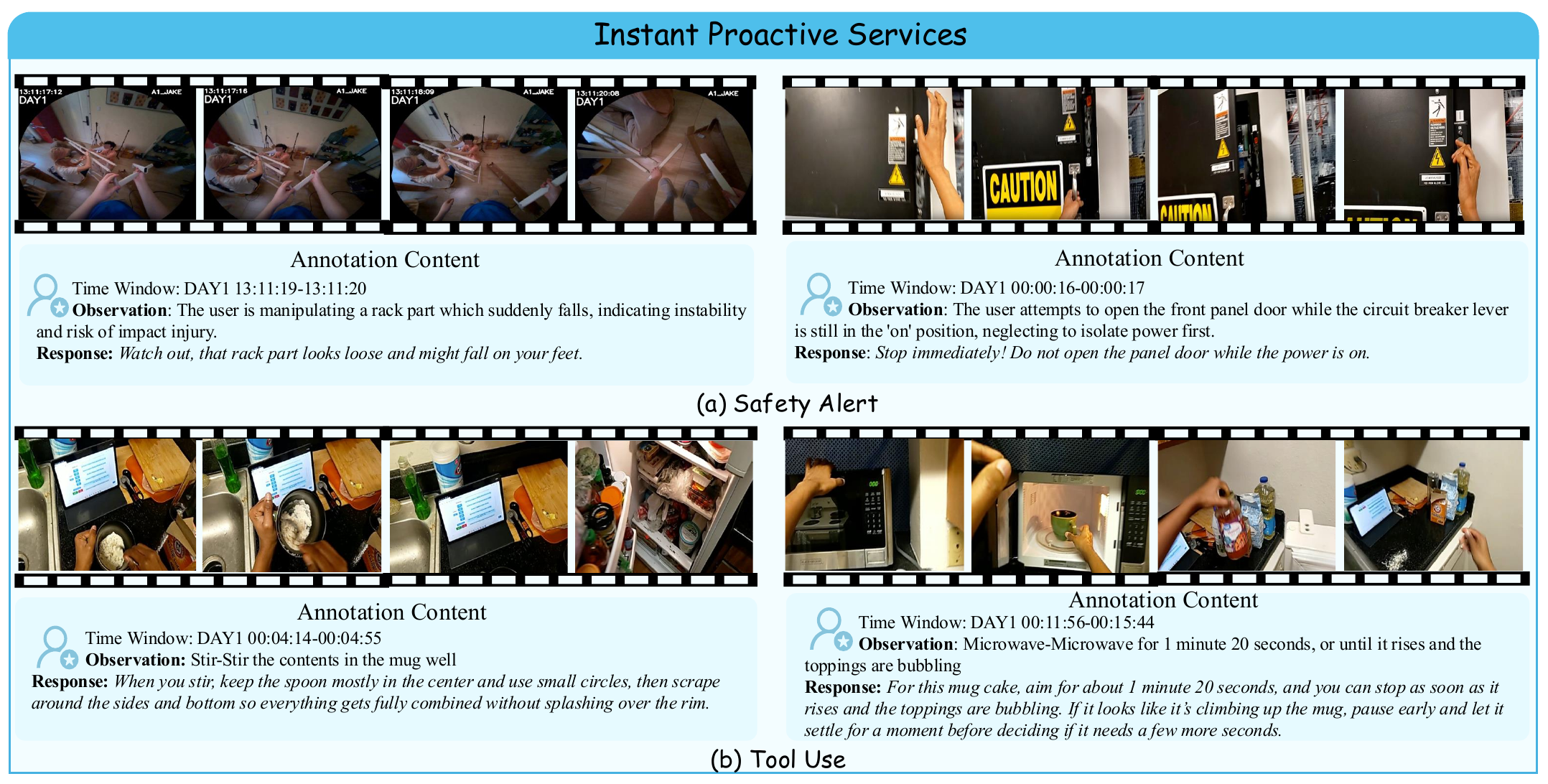}
    \caption{Annotation examples of Instant proactive assistance.}
    \label{fig:instant_vis}
\end{figure}

\begin{figure}
    \centering
    \includegraphics[width=1\linewidth]{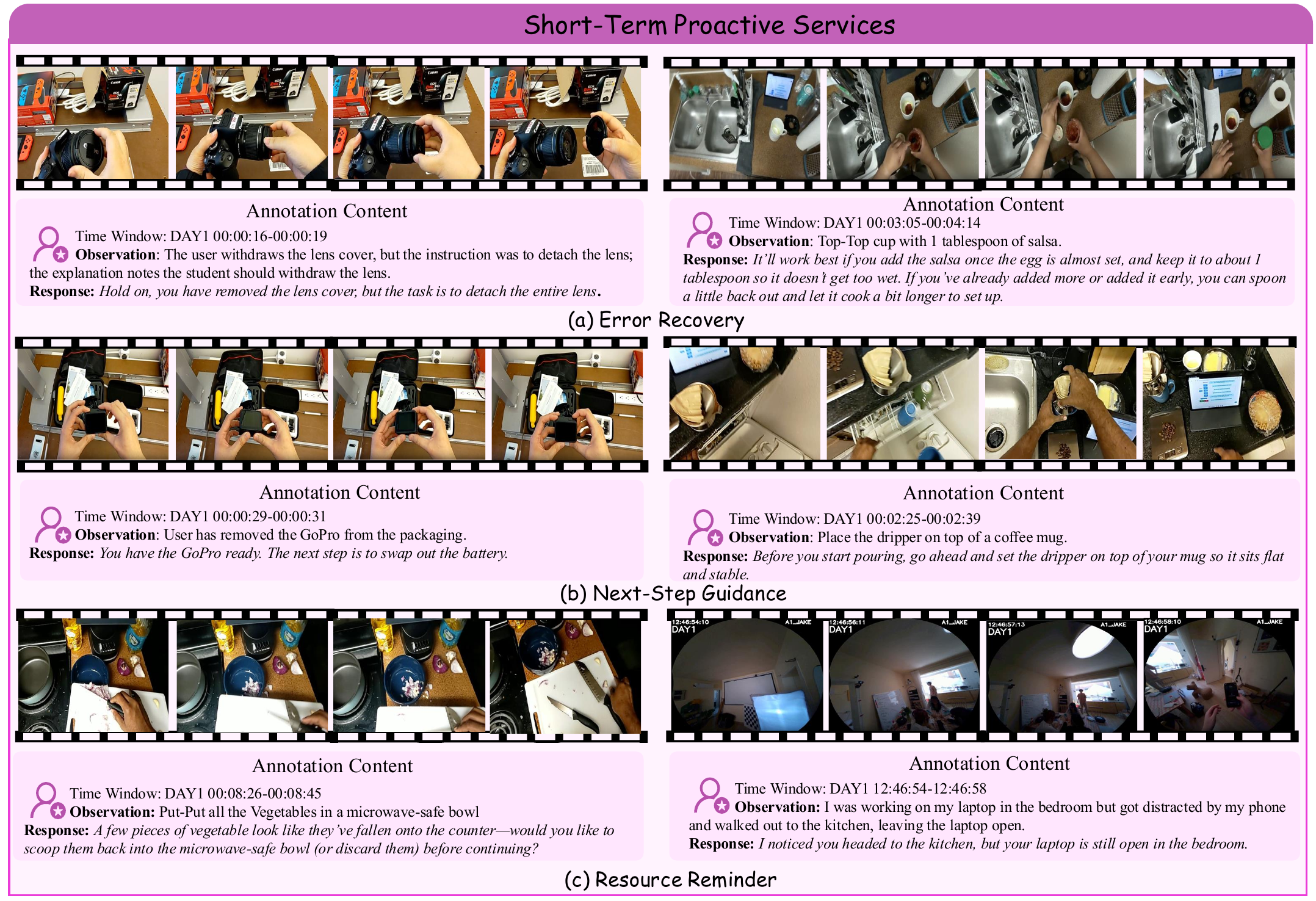}
    \caption{Annotation examples of Short-term proactive assistance.}
    \label{fig:short_term_vis}
\end{figure}

\begin{figure}
    \centering
    \includegraphics[width=1\linewidth]{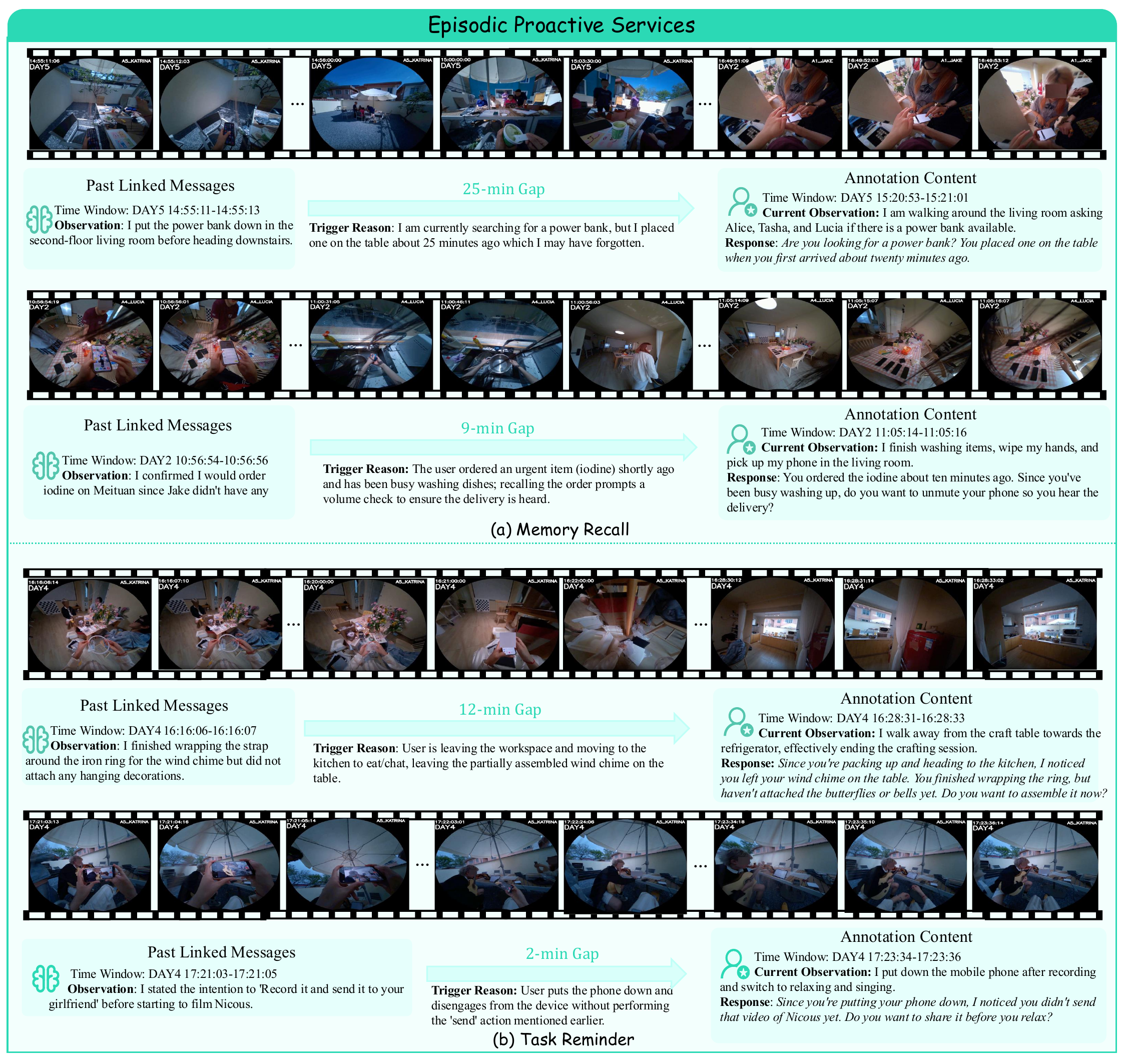}
    \caption{Annotation examples of Episodic proactive assistance.}
    \label{fig:episodic_vis}
\end{figure}

\begin{figure*}
    \centering
    \includegraphics[width=0.85\linewidth]{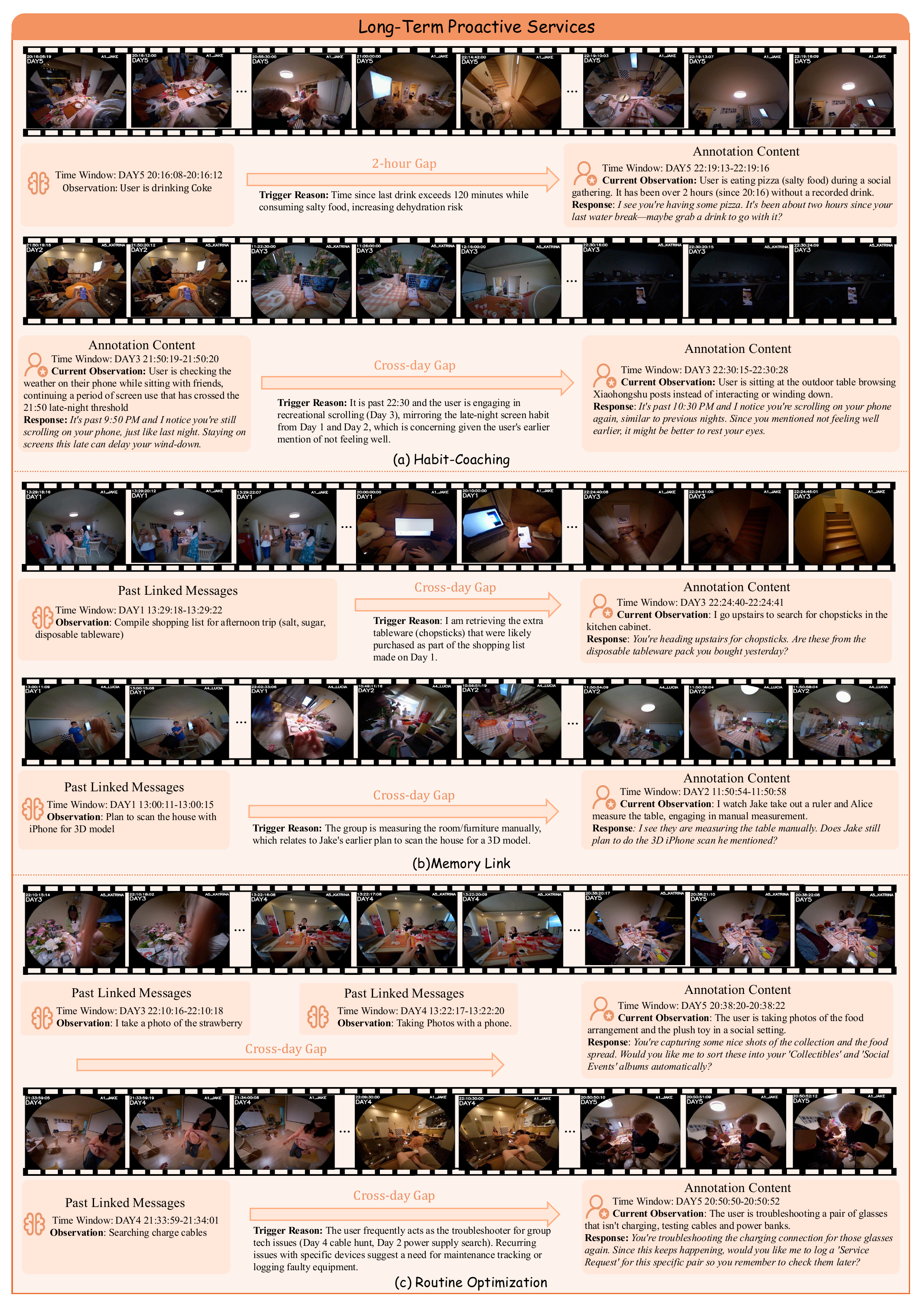}
    \caption{Annotation examples of Long-term proactive assistance.}
    \label{fig:long_term_vis}
\end{figure*}



%
%
\bibliographystyle{splncs04}
\bibliography{main}
\end{document}